\documentclass[11pt,a4paper]{article}

\usepackage[margin=0.85in]{geometry}
\usepackage{amsmath,amssymb,mathtools,bm}
\usepackage{booktabs,tabularx,array,longtable}
\usepackage{enumitem}
\usepackage{setspace}
\usepackage{titlesec}
\usepackage{titling}
\usepackage{natbib}
\usepackage{hyperref}
\usepackage{xurl}
\usepackage{microtype}
\usepackage{graphicx}
\usepackage{caption}
\usepackage{xcolor}
\usepackage{float}
\usepackage{placeins}
\usepackage{tikz}
\usetikzlibrary{arrows.meta,positioning}

\hypersetup{
  colorlinks=true,
  linkcolor=black,
  citecolor=black,
  urlcolor=blue,
  pdftitle={The Calibration Turn in AI-Assisted Research: A Conceptual and Methodological Framework for Evidence-Licensed Claims},
  pdfauthor={Hongmin Li}
}
\setlist[itemize]{leftmargin=1.8em,itemsep=0.08em,topsep=0.18em,parsep=0pt}
\setlist[enumerate]{leftmargin=1.8em,itemsep=0.08em,topsep=0.18em,parsep=0pt}
\titlespacing*{\section}{0pt}{1.15ex plus .35ex minus .2ex}{0.65ex plus .2ex}
\titlespacing*{\subsection}{0pt}{0.95ex plus .3ex minus .2ex}{0.45ex plus .15ex}
\titlespacing*{\subsubsection}{0pt}{0.8ex plus .25ex minus .15ex}{0.35ex plus .1ex}
\pretitle{\begin{center}\Large\bfseries}
\posttitle{\par\end{center}\vspace{-0.55em}}
\preauthor{\begin{center}\normalsize}
\postauthor{\par\end{center}\vspace{-0.85em}}
\predate{\begin{center}\small}
\postdate{\par\end{center}\vspace{-1.1em}}

\newcommand{\E}{\mathbb{E}}
\newcommand{\doop}{\operatorname{do}}
\newcommand{\Sem}{\operatorname{Sem}}
\newcommand{\Conseq}{\operatorname{Conseq}}
\newcommand{\denote}[1]{\left[\!\left[#1\right]\!\right]}
\newcommand{\argmax}{\operatorname*{arg\,max}}
\newcommand{\argmin}{\operatorname*{arg\,min}}
\newcommand{\Claim}{\operatorname{Claim}}
\newcommand{\Strength}{\operatorname{Strength}}
\newcommand{\Know}{\mathcal{K}}
\newcommand{\Cal}{\operatorname{Cal}}
\newcommand{\Req}{\operatorname{Req}}
\newcommand{\Sup}{\operatorname{Sup}}
\newcommand{\Cost}{\operatorname{Cost}}
\newcommand{\Risk}{\operatorname{Risk}}
\newcommand{\Debt}{\operatorname{Debt}}
\newcommand{\Shear}{\operatorname{Shear}}
\newcommand{\Max}{\operatorname{Max}}
\newcommand{\Dom}{\mathsf{D}}
\newcolumntype{Y}{>{\raggedright\arraybackslash}X}

\title{\textbf{The Calibration Turn in AI-Assisted Research}\\[0.25em]
\large A Conceptual and Methodological Framework for Evidence-Licensed Claims}
\author{Hongmin Li\textsuperscript{1,2}\thanks{Author note: Hongmin Li is a Researcher at the School of Life Science and Technology, Institute of Science Tokyo, and a Guest Researcher at the Department of Computational Biology and Medical Sciences, Graduate School of Frontier Sciences, The University of Tokyo. ORCID: \href{https://orcid.org/0000-0003-0228-0600}{0000-0003-0228-0600}. Correspondence: \texttt{lihongmin@edu.k.u-tokyo.ac.jp}.}\\[0.45em]
\textsuperscript{1}School of Life Science and Technology, Institute of Science Tokyo,\\
2-12-1 Ookayama, Meguro-ku, Tokyo 152-8550, Japan\\[0.35em]
\textsuperscript{2}Department of Computational Biology and Medical Sciences,\\
Graduate School of Frontier Sciences, The University of Tokyo,\\
5-1-5 Kashiwanoha, Kashiwa-shi, Chiba 277-8561, Japan}
\date{June 16, 2026}

\begin{document}
\maketitle

\begin{abstract}
AI-assisted research has entered a stage in which the central question is no longer only whether AI systems can generate hypotheses, run experiments, or produce manuscripts, but whether their final scientific claims are calibrated to the evidence that supports them. This Perspective-style paper develops a conceptual and methodological framework for evaluating evidence-licensed claims in AI-assisted research. The framework is motivated by a methodological comparison of representative routes, including specialized scientific foundation models, human-in-the-loop LLM research assistants, multi-agent co-scientist systems, end-to-end AI Scientist pipelines, algorithmic and mathematical discovery agents, and self-driving laboratories. It represents AI-assisted research as a composition of five operators: hypothesis generation, model-mediated consequence derivation, external validation, belief update, and claim calibration. The central claim is that calibration is not merely cautious wording. It is a mechanism for managing scientific assertion rights: evidence licenses some forms of speech and withholds others. The paper distinguishes linguistic, consequence-based, interventional, and evidence-licensed semantics; defines the claim-evidence gap and epistemic debt; and treats minimal structural reconstruction across heterogeneous outputs as a special, upward form of claim calibration. It argues that AI science routes differ not only in consequence-derivation capacity or automation, but also in whether their evaluators are independent, reliable, world-grounded, and translated into bounded scientific claims. AISim-Cal is included only as an illustrative synthetic dynamics exercise, not as an empirical forecast or benchmark. The resulting calibration principles are: no claim without license, validation does not determine claim level, and automation amplifies the need for calibration. This framework therefore evaluates reliable AI-assisted research as an AI-augmented loop that generates hypotheses, derives testable consequences, accepts independent adjudication, updates beliefs, and outputs only evidence-licensed claims.

\end{abstract}

\section{Introduction: From Scientific Automation to Claim Licensing}

Discussions of AI scientific exploration often focus on whether a system has produced a sufficiently substantive scientific contribution or a genuine discovery. This issue is broader than model capability or manuscript generation, but it remains incomplete unless contribution and discovery are constrained by evidence. Scientific knowledge is not primarily a matter of novelty labels or textual productivity; it is a matter of testable constraint on the world. A chapter by King and Zenil in the OECD report on artificial intelligence in science frames the goal of science as the construction of models that predict what happens in the real world, especially in experiments, and treats predictive performance on experiments as a natural objective for AI systems in science \citep{OECD2023Framework,KingZenil2023AIDrivenScience}. This paper adopts that predictive framing where it applies, but treats prediction as an important special case of a broader consequence relation. Some sciences derive future-oriented predictions; others derive retrodictive traces, diagnostic features, proof obligations, classification constraints, or measurement signatures. This framing is also continuous with older ideas about pragmatic meaning and inquiry, falsifiability, research programmes, intervention, severe testing, and the relation between claims, data, and warrants \citep{Peirce1878Ideas,Popper1959Logic,Lakatos1970Falsification,Hacking1983Representing,Mayo1996Error,Toulmin1958Uses}. This provides the starting point for the present paper: an AI scientific system is reliable only insofar as it can turn hypotheses into domain-appropriate testable consequences, update those hypotheses through independent adjudication, and calibrate its final claims to the evidence produced by that adjudication.

As of June 2026, this Perspective uses six non-exhaustive route families as epistemological comparators. AlphaFold 3 \citep{Abramson2024AlphaFold3} and GNoME \citep{Merchant2023GNoME} represent specialized scientific foundation models. AlphaFold 3 uses a substantially updated diffusion-based architecture to predict the joint structure of biomolecular complexes that may include proteins, nucleic acids, small molecules, ions, and modified residues \citep{Abramson2024AlphaFold3}. GNoME uses deep learning to expand the search space for materials and generate large-scale computational candidates for stable crystal structures \citep{Merchant2023GNoME}. GPT-5 and related frontier models are used by experts for literature synthesis, proof sketches, mechanistic hypotheses, and experimental suggestions, but the published case reports explicitly describe curated examples rather than systematic samples and emphasize the need for expert validation \citep{Bubeck2025GPT5Science,OpenAI2025ScienceGPT5}. Google Co-Scientist and FutureHouse Robin organize hypothesis generation, critique, ranking, data analysis, and experimental planning into multi-agent systems \citep{Gottweis2026CoScientist,Ghareeb2026Robin}. The Sakana AI Scientist line attempts to automate idea generation, code execution, figures, manuscripts, and review in machine learning research \citep{Lu2026AIScientist}. AlphaEvolve, AlphaProof, and AlphaGeometry 2 show that when candidate objects can be evaluated by program tests, benchmarks, or formal proof checkers, AI systems can obtain much stronger epistemic warrants \citep{Novikov2025AlphaEvolve,Hubert2026AlphaProof}. Boiko et al.'s Coscientist chemistry system and A-Lab connect AI systems to robotic experimentation and real data feedback, while also exposing the high cost of experimental engineering, interpretation, and reproducibility review \citep{Boiko2023Coscientist,Szymanski2023ALab,Chawla2026ALabCorrection}. The evidential status of these sources differs: peer-reviewed articles, arXiv or white-paper claims, official reports, and journalism-supported corrections are treated here as epistemological comparators, not as independently verified field rankings.

The central argument of this paper is that these routes differ less in their model families than in their implicit theory of scientific meaning. Some systems locate meaning in prediction distributions, some in textual hypotheses, some in research artifacts, some in programs or proofs, and some in experimental responses. Across routes, the shared failure mode is claim drift: the migration from a licensed weaker claim to an unlicensed stronger one. The key issue is therefore not only the identity and independence of the adjudicator, but also the strength of scientific assertion licensed after adjudication.

The paper addresses three questions. First, how does an AI-generated hypothesis become a scientific claim? Second, what level of claim is licensed by a given external validation result? Third, when evidence is insufficient, how should the system cancel the resulting epistemic debt by collecting more evidence, weakening the claim, narrowing its boundary, or changing its modal status?

The contributions are fourfold. First, the paper introduces an evidential calibration vocabulary for AI scientific exploration, moving the analysis from hypothesis generation and external validation to evidence-licensed assertion. Second, it formalizes scientific assertion rights at a schematic level through a domain-sensitive license relation $E\Vdash_{\Dom,V} C$, together with claim orders and claim-relative evidence preorders. Third, it defines the claim-evidence gap and epistemic debt as diagnostic measures of mismatch between the evidential strength required by a claim and the support actually provided by the evidence. Fourth, it uses representative AI science routes as methodological comparators to argue that the central risk is not simply lack of validation, but the overinterpretation of experimental actionability, artifact completeness, or paper form as stronger scientific discovery than the evidence permits.

This paper develops the methodological and epistemological framework. It does not introduce a new AI-assisted science system or report domain-specific scientific results.

This is a Perspective-style theoretical synthesis and methodological framework rather than a systematic literature review. The routes discussed below are selected as prominent epistemological comparators for AI-assisted research, not as an exhaustive census of all systems.

In this paper, ``AI-assisted research'' is the umbrella term, while ``AI scientific exploration'' refers to the subset of AI-assisted research systems that generate, evaluate, or communicate scientific hypotheses and claims.

\textbf{Article type and scope.} The primary contribution of this paper is a conceptual and methodological framework for calibrating scientific claims in AI-assisted research. The route comparison motivates and stress-tests the framework across representative paradigms; it is not a systematic field map. AISim-Cal is an illustrative synthetic dynamics exercise; it is not an empirical forecast, a benchmark, or evidence that any route is superior. The phrase ``calibration turn'' names the evaluative shift proposed here, not a claim that the field has already converged on a settled paradigm.

Every calibrated scientific output should therefore specify both the claim licensed by the evidence and the stronger paired non-claim that remains unlicensed.

\begin{center}
\fbox{\begin{minipage}{0.92\textwidth}
\textbf{Core thesis.} AI scientific exploration should be evaluated not only by whether a system generates hypotheses or obtains validation, but by whether it manages scientific assertion rights. We formalize this with the license relation $E\Vdash_{\Dom,V} C$, the claim-evidence gap $\Delta_{\Dom,V}(C,E)$, the epistemic debt $\Debt^\ell_{\Dom,V}(C,E)$, and the calibration operator $\Cal_{\Dom}$, which maps a raw claim, evidence, evaluator, and domain context to a maximal evidence-licensed claim frontier. A selected calibrated claim is then $C^\ast\in\Cal_{\Dom}(C,E,V)$. Experimental actionability, artifact completeness, or benchmark success becomes scientifically reliable only when it is converted into an evidence-licensed claim.
\end{minipage}}
\end{center}

The framework is designed to connect philosophical semantics with AI-system analysis. Definitions are repeated when they become operationally relevant, and key symbols are explained both in the notation table and near the formulas in which they are used. This redundancy supports comparisons across LLM-based scientific agents, AI Scientist pipelines, autonomous laboratories, formal proof agents, and specialized scientific models. In each case, the same decomposition is required: which subsystem generates hypotheses, which subsystem derives consequences, which mechanism functions as an evaluator, and where the system's language must be downgraded when evidence is insufficient.

The same question governs all routes considered below: what claim does the evidence license? This question remains invariant across language models, formal proof systems, robotic laboratories, benchmark-driven machine learning, and domain-specific foundation models, even though their artifacts, evaluators, costs, and validation schedules differ substantially.

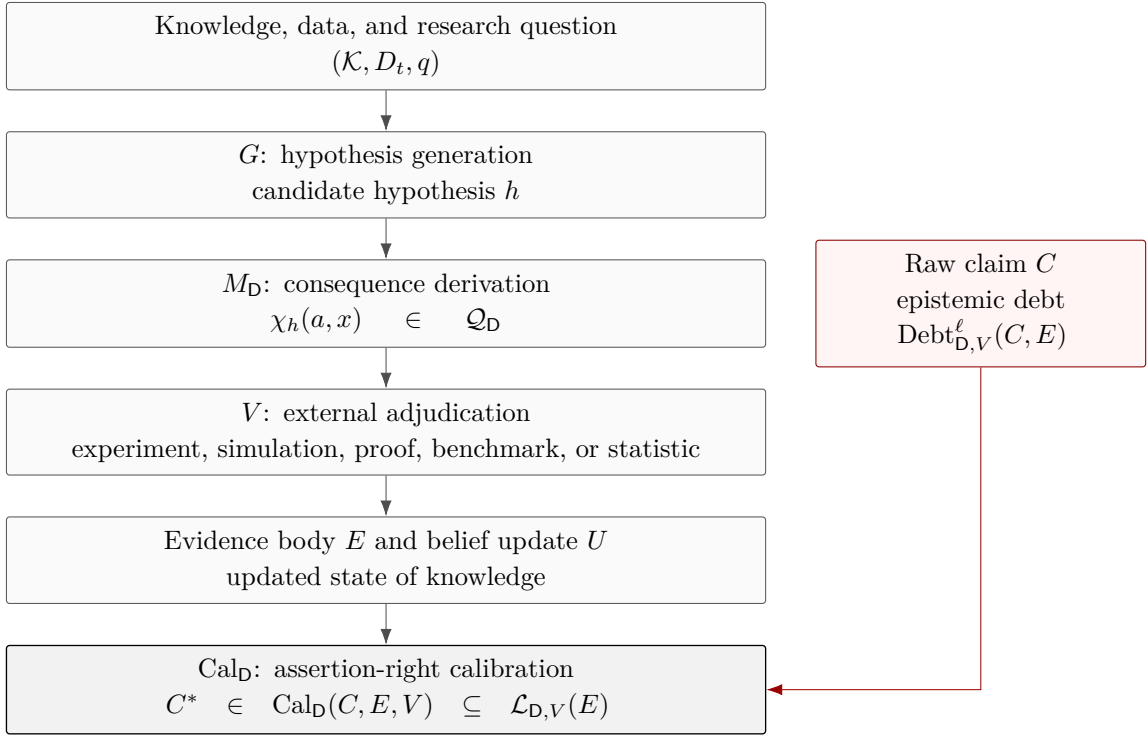
\begin{figure}[H]
\centering
\small
\begin{tikzpicture}[
  font=\small,
  node distance=0.54cm,
  flowbox/.style={
    draw=black!65,
    fill=black!2,
    rounded corners=1.2pt,
    align=center,
    text width=0.58\textwidth,
    minimum height=1.02cm,
    inner sep=5pt
  },
  calbox/.style={
    flowbox,
    draw=black,
    fill=black!5,
    line width=0.5pt
  },
  sidebox/.style={
    draw=red!55!black,
    fill=red!4,
    rounded corners=1.2pt,
    align=center,
    text width=0.24\textwidth,
    inner sep=5pt
  },
  arrow/.style={-{Latex[length=2.1mm,width=1.5mm]},line width=0.38pt,draw=black!70},
  riskarrow/.style={-{Latex[length=2.1mm,width=1.5mm]},line width=0.38pt,draw=red!60!black}
]
\node[flowbox] (input) {Knowledge, data, and research question\\$(\Know,D_t,q)$};
\node[flowbox, below=of input] (gen) {$G$: hypothesis generation\\candidate hypothesis $h$};
\node[flowbox, below=of gen] (model) {$M_{\Dom}$: consequence derivation\\$\chi_h(a,x)\in\mathcal{Q}_{\Dom}$};
\node[flowbox, below=of model] (valid) {$V$: external adjudication\\experiment, simulation, proof, benchmark, or statistic};
\node[flowbox, below=of valid] (evid) {Evidence body $E$ and belief update $U$\\updated state of knowledge};
\node[calbox, below=of evid] (cal) {$\Cal_{\Dom}$: assertion-right calibration\\$C^\ast\in\Cal_{\Dom}(C,E,V)\subseteq\mathcal{L}_{\Dom,V}(E)$};
\node[sidebox, right=0.65cm of model] (raw) {Raw claim $C$\\epistemic debt\\$\Debt^\ell_{\Dom,V}(C,E)$};

\draw[arrow] (input) -- (gen);
\draw[arrow] (gen) -- (model);
\draw[arrow] (model) -- (valid);
\draw[arrow] (valid) -- (evid);
\draw[arrow] (evid) -- (cal);
\draw[riskarrow] (raw.south) |- (cal.east);
\end{tikzpicture}
\caption{The calibrated AI science loop. Hypothesis generation $G$ proposes candidates from knowledge, accumulated data $D_t$, and a research question. Modeling $M_{\Dom}$ turns a hypothesis into domain-appropriate testable consequences. Interventional prediction $P_h(Y\mid \doop(a),x)$ is one important special case; other domains may derive retrodictive traces, diagnostic features, proof obligations, classification constraints, or measurement signatures. Validation $V$ adjudicates the consequence through experiment, simulation, proof checking, benchmarking, measurement, or statistical testing. Belief update $U$ incorporates the result into the current state of knowledge. Claim calibration $\Cal_{\Dom}$ maps the raw claim $C$, evidence $E$, evaluator $V$, and domain context $\Dom$ to a maximal licensed frontier, from which a selected calibrated claim $C^\ast$ must belong to $\mathcal{L}_{\Dom,V}(E)$. Without $\Cal_{\Dom}$, an AI science loop can obtain validation while still producing overstrong claims and epistemic debt.}
\label{fig:calibrated-loop}
\end{figure}

\begin{table}[H]
\centering
\scriptsize
\caption{Licensed claims and paired non-claims across representative AI science routes}
\label{tab:paired-nonclaims}
\begin{tabularx}{\textwidth}{@{}>{\raggedright\arraybackslash}p{2.35cm}>{\raggedright\arraybackslash}p{2.55cm}>{\raggedright\arraybackslash}p{2.55cm}Y Y@{}}
\toprule
Route & Raw output & Evidence object & Licensed claim & Paired non-claim \\
\midrule
Specialized foundation model & Predicted structure, property, or candidate & Model prediction, confidence, structural or downstream validation & Candidate is supported by the specified predictive model or validation setting. & Not experimental proof of mechanism, function, or broad biological/material discovery. \\
Multi-agent co-scientist & Mechanistic, drug-repurposing, or experimental hypothesis & Literature-grounded rationale, ranking, selected assays, or data analysis & Candidate hypothesis or combination is supported under the reported search and assay conditions. & Not clinically validated therapy or generally established biological mechanism. \\
End-to-end AI Scientist & Idea, code, experiment, figures, and manuscript artifact & Benchmark result, execution trace, review score, and reproducibility package & The workflow automates substantial research-artifact generation under constrained computational settings. & Not general autonomous scientific discovery across domains. \\
Algorithmic/proof agent & Program, algorithm, conjecture, proof sketch, or certificate & Unit test, benchmark, proof checker, compiler, or formal verifier & Candidate satisfies the stated formal, programmatic, or benchmark constraints. & Not automatically broad empirical science or mechanism-level explanation. \\
Self-driving laboratory & Experiment plan, synthesized target, or optimization trace & Robotic execution, assay readout, characterization, and replication record & The workflow realized or optimized targets under specified experimental and characterization criteria. & Not automatically novelty, general mechanism, or application-ready material/chemical discovery. \\
\bottomrule
\end{tabularx}
\end{table}

\section{Notation and Core Symbols}

Because the paper compares several AI science routes, the same mathematical section contains variables for worlds, hypotheses, claims, evaluators, and workflow artifacts. Table~\ref{tab:notation} gives a consolidated glossary: the left column gives the symbol, the middle column gives its formal role, and the right column gives the plain interpretation used in the argument.

Three conventions reduce symbol overload. First, uppercase $C$ is reserved for scientific claims in expressions such as $\Claim(h,s,b,m)$, $\Delta_{\Dom,V}(C,E)$, or $\Cal_{\Dom}(C,E,V)$. Costs are written as $\Cost(\cdot)$, not $C(\cdot)$. Second, accumulated data are written as $D_t$, whereas the domain context is written as $\Dom$ when it indexes a calibration or licensing relation. Third, claim requirements and evidential support are written as $\Req_{\Dom}(C)$ and $\Sup_{\Dom,V}(E,C)$, not as generic $R$ and $S$. This separation is important because $\Cal_{\Dom}$ is the paper's central operator.

The most important symbols for the whole argument are $h$, $C$, $E$, $V$, and $\Cal_{\Dom}$. The hypothesis $h$ is what an AI system proposes or models. The claim $C$ is what the system, paper, or user-facing report says about that hypothesis. The evidence $E$ is what has actually been observed, tested, proven, simulated, or replicated. The evaluator $V$ is the mechanism that judges whether the evidence supports the candidate. The calibration operator $\Cal_{\Dom}$ converts the raw claim into a bounded claim $C^\ast$ that does not exceed the evidence. The core dependency is:
\[
\boxed{\text{hypothesis }h \longrightarrow \text{claim }C \longrightarrow \text{evidence }E \longrightarrow \text{calibrated claim }C^\ast.}
\]

{\small
\begin{longtable}{@{}p{2.8cm}p{4.1cm}p{6.4cm}@{}}
\caption{Main notation used in the manuscript}
\label{tab:notation}\\
\toprule
Symbol & Formal role & Plain interpretation in this paper \\
\midrule
\endfirsthead
\toprule
Symbol & Formal role & Plain interpretation in this paper \\
\midrule
\endhead
\bottomrule
\endfoot
$\Omega$ & Space of possible worlds or mechanisms & The set of causal, physical, biological, material, or mathematical structures that could explain the data. \\
$\omega^\ast$ & True but unknown element of $\Omega$ & The real mechanism that science tries to approximate but cannot observe directly. \\
$D_t$ & Data accumulated up to time $t$ & Everything observed so far: literature, experiments, simulations, benchmark results, proof attempts, and database records. \\
$\Dom$ & Domain context & The field-specific context that determines admissible evidence, claim strength, and calibration norms. \\
$(a_i,y_i,c_i)$ & Action--outcome--context triple & The $i$th test or query: what was done, what happened, and under which conditions. \\
$a,a_t$ & Scientific action or test & An experiment, intervention, simulation, query, proof attempt, benchmark run, or robotic laboratory action. \\
$\mathcal{A}$ & Action space & The set of interventions, experiments, simulations, proof attempts, or queries available to the system. \\
$\mathcal{X}$ & Context space & The set of biological, chemical, computational, or experimental contexts; this avoids using $C$ for contexts. \\
$y,Y$ & Observed or random outcome & The measured response produced by an action; $Y$ is the random variable and $y$ is a realized observation. \\
$c$ & Context condition & The biological, chemical, computational, or experimental setting in which an action is evaluated. \\
$\Know$ & Prior knowledge base & Literature, databases, theory, expert background, and previously accepted constraints. \\
$q$ & Research question & The problem statement or scientific task that guides hypothesis generation. \\
$h$ & Scientific hypothesis & A proposed mechanism, relation, conjecture, explanation, or candidate scientific statement before claim calibration. \\
$H_t$ & Hypothesis set at time $t$ & The current population of hypotheses available to the AI or scientist. \\
$M_t$ & Current model & The predictive, mechanistic, statistical, or simulation model currently used to derive consequences. \\
$B_t(\omega)$ & Belief distribution over $\Omega$ & The current degree of belief that mechanism $\omega$ is the correct one after seeing $D_t$. \\
$b_t(\omega)$ & Belief state in the SDL section & Same idea as $B_t(\omega)$, used in the self-driving-laboratory active-learning notation. \\
$S_t$ & Inquiry state & The full state $(D_t,\Know,H_t,M_t,B_t)$ used by a scientific policy. \\
$\pi$ & Scientific policy & A rule that chooses the next action $a_{t+1}$ from the current state $S_t$. \\
$\Phi(a_t,S_t)$ & Local scientific utility & The immediate value of action $a_t$, combining prediction, causality, explanation, novelty, cost, and risk. \\
$\Delta\mathrm{Pred}$ & Predictive improvement & How much better the system predicts observations after a proposed action or update. \\
$\Delta\mathrm{Causal}$ & Causal-identification improvement & How much the action clarifies intervention-relevant causal structure. \\
$\Delta\mathrm{Explanation}$ & Explanatory improvement & How much the action improves mechanism-level interpretation. \\
$\Delta\mathrm{Novelty}$ & Novelty improvement & How much the action expands beyond known hypotheses or artifacts. \\
$\Cost(a_t)$ & Cost of action $a_t$ & A local cost term, kept distinct from the scientific claim $C$. \\
$\Risk(a_t)$ & Risk of action $a_t$ & A local safety or feasibility penalty, kept distinct from the requirement profile $\Req_{\Dom}(C)$. \\
$\denote{h}$ & Model-theoretic denotation & The set of possible worlds in which hypothesis $h$ is true. \\
$\Sem(h)$ & Scientific semantics of $h$ & The domain-appropriate consequences that make a hypothesis scientifically testable. \\
$\Conseq_{\Dom}(h)$ & Consequence profile under $h$ & The set of testable consequences that hypothesis $h$ implies in domain $\Dom$. \\
$\chi_h(a,x)$ & Local consequence under $h$ & A predicted outcome, retrodictive trace, proof obligation, diagnostic feature, measurement signature, or constraint induced by $h$. \\
$\mathcal{Q}_{\Dom}$ & Domain consequence space & The space containing the kinds of consequences admissible in domain $\Dom$. \\
$P_h(Y\mid \doop(a),x)$ & Interventional prediction under $h$ & The special case in which the consequence is a probability distribution over outcomes after action $a$ in context $x$. \\
$\doop(a)$ & Intervention operator & The act of doing or enforcing action $a$, rather than merely observing correlation. \\
$\mathcal{T}(h)$ & Test set for $h$ & The experiments, simulations, benchmarks, or proof obligations that can evaluate $h$. \\
$\mathcal{V}(h)$ & Verification or refutation rule & The rule that says what would count as support, failure, or refutation for hypothesis $h$. \\
$\mathcal{B}(h)$ & Claim boundary for $h$ & The range of assertions about $h$ that current evidence is allowed to support. \\
$C$ & Scientific claim & A sentence-level assertion made in a paper, report, or system output. \\
$\mathcal{C}_{\Dom}$ & Domain-specific claim space & The set of claims that can be made in domain context $\Dom$. \\
$\preceq_{\Dom}$ & Claim-strength order & $C_1\preceq_{\Dom} C_2$ means that $C_1$ is no stronger than $C_2$ in domain $\Dom$. \\
$\Claim(h,s,b,m)$ & Structured claim representation & A claim built from a hypothesis $h$, strength $s$, boundary $b$, and modal status $m$. \\
$s$ & Claim strength & How strong the assertion is: possible, supported, demonstrated, general, translational, and so on. \\
$b$ & Claim boundary & The domain, population, assay, benchmark, material class, or context where the claim is asserted to hold. \\
$m$ & Modal status & The claim's epistemic mode: possible, supported, demonstrated, explained, discovered, or translated. \\
$E$ & Evidence body & The experiments, simulations, proofs, benchmarks, databases, or replications available for a claim. \\
$\mathcal{E}_{\Dom}$ & Domain-specific evidence space & The set of evidence bodies admissible in domain $\Dom$. \\
$\mathcal{W}_{\Dom}$ & Warrant-profile space & The ordered space in which evidential requirements and support profiles are compared. \\
$E_1\sqsubseteq_{\Dom,V,\mathcal{C}_0} E_2$ & Claim-relative evidence order & Evidence $E_2$ is no weaker than $E_1$ for every claim in the comparison family $\mathcal{C}_0$ under evaluator $V$. \\
$E\Vdash_{\Dom,V} C$ & License relation & Under domain $\Dom$ and evaluator $V$, evidence $E$ licenses scientific assertion $C$; this is warrant, not truth. \\
$\Req_{\Dom}(C)$ & Requirement profile for claim $C$ & The evidential warrant needed before claim $C$ is legitimate in domain context $\Dom$. \\
$\Sup_{\Dom,V}(E,C)$ & Support profile provided by $E$ for $C$ & The evidential support that the available evidence and evaluator give to the claim. \\
$\Sup_{\Dom,V}\succeq_{\mathcal{W},\Dom}\Req_{\Dom}$ & Evidence-licensing relation & The support profile $\Sup_{\Dom,V}(E,C)$ covers the requirement profile $\Req_{\Dom}(C)$. \\
$\mathcal{F}$ & Family of heterogeneous claims & A comparison family $\{C_1,\ldots,C_k\}$ whose members may share a higher-order structure. \\
$C_{\mathrm{str}}$ & Candidate structural claim & A proposed family-level minimal reconstruction that captures a candidate common structure across $\mathcal{F}$; it is still a raw claim until calibrated. \\
$E_{\mathcal{F}},V_{\mathcal{F}}$ & Family-level evidence and evaluator record & The evidence and evaluator provenance used to decide whether a structural reconstruction is licensed. \\
$\ell_{\Dom}$ & Scalarization map & An order-preserving map from warrant profiles into the scalar diagnostic $\Delta_{\Dom,V}(C,E)$. \\
$\Delta_{\Dom,V}(C,E)$ & Domain-indexed claim-evidence gap & A diagnostic projection of $\Req_{\Dom}(C)$ minus $\Sup_{\Dom,V}(E,C)$; positive values mean the claim is stronger than the evidence under domain context $\Dom$. \\
$\Debt^\ell_{\Dom,V}(C,E)$ & Scalar epistemic debt & The positive part of the scalarized claim-evidence gap, or the unmet warrant owed by an overstrong claim under scalarization $\ell_{\Dom}$. \\
$\mathcal{L}_{\Dom,V}(E)$ & Licensed claim set & All claims licensed by evidence $E$ under domain context $\Dom$ and evaluator $V$; a lower set in the claim poset. \\
$\bot_{\Dom}$ & Bottom claim & A fallback claim stating that no scientific assertion beyond proposal is licensed. \\
$\downarrow C$ & Down-set of a claim & All claims no stronger than $C$: $\{C':C'\preceq_{\Dom} C\}$. \\
$C'\preceq_{\Dom} C$ & Claim weakening relation & Claim $C'$ is no stronger than the raw or intended claim $C$ in domain context $\Dom$. \\
$\Strength(C')$ & Strength score of a claim & A ranking of how strong or informative a licensed claim is. \\
$C^\ast$ & Selected calibrated claim & A reportable claim selected from the maximal evidence-licensed frontier $\Cal_{\Dom}(C,E,V)$. \\
$\Cal_{\Dom}$ & Claim-calibration operator & The assertion-right operator: it maps a raw claim, evidence, evaluator, and domain context to a maximal licensed claim frontier; a domain-specific selection rule may choose $C^\ast$ from this frontier. \\
$C_{\mathrm{literal}}$ & Literal claim & The exact sentence or formal claim written in a paper, report, title, or system output. \\
$C_{\mathrm{implied}}$ & Implied claim & The stronger or broader claim induced by framing, title, abstract, figure, press release, or downstream communication. \\
$\Shear_r(t)$ & Epistemic shear & A route-level imbalance between generation/consequence-derivation capacity and adjudication/calibration capacity. \\
$Q^{\mathrm{cal}}_{r,\Dom}(t)$ & Route-level calibration quality & A scalar AISim-Cal simulation parameter for how well route $r$ pulls claims back toward the licensed frontier in domain $\Dom$; kept distinct from the operator $\Cal_{\Dom}$. \\
$\Gamma_{\Dom}$ & Goal clarity in domain $\Dom$ & How well a domain or task has specified its objective, scoring rule, and claim-value target. \\
$\tau_\Gamma$ & Goal-clarity threshold & The minimum clarity level used in AISim-Cal before a route ordering is reported. \\
$A_{r,\Dom}$ & Route-domain applicability & How compatible route $r$ is with domain $\Dom$; it prevents strong evaluators from implying universal route superiority. \\
$d,i,r,e,c_{\mathrm{id}},\xi,n$ & Support dimensions & Directness, independence, reproducibility, effect size, causal identification, external validity, and negative evidence. \\
$F(\cdot)$ & Support aggregation or profiling function & The domain-specific rule that combines evidential dimensions into $\Sup_{\Dom,V}(E,C)$. \\
$L0$--$L6$ & Claim ladder levels & A practical scale from hypothesis generation to translational or application-ready assertion. \\
$G$ & Hypothesis-generation operator & Maps knowledge, data, and question to a hypothesis set. \\
$M$ & Model-mediated consequence-derivation operator & Maps hypothesis, action, and context to a domain-appropriate consequence profile. \\
$V$ & Validation or evaluator operator & Judges whether a hypothesis, consequence, program, proof, or experiment survived a test. \\
$U$ & Belief-update operator & Updates the belief state after new evidence is observed. \\
$W(h,C_h,E_h)$ & Discovery objective & A score for hypotheses and intended claims that rewards evidence and penalizes overclaiming. \\
$W_{\mathrm{text}}$ & Textual warrant & Textual plausibility or continuation-based support. \\
$W_{\mathrm{conseq}}$ & Consequence warrant & Evidence that a hypothesis yields domain-appropriate testable consequences. \\
$W_{\mathrm{adjudicated}}$ & Adjudicated warrant & Evidence that derived consequences have survived an external evaluator, such as a test, proof checker, benchmark, or assay. \\
$W_{\mathrm{license}}$ & Evidence-licensed assertion & The strongest level here: a claim is inside $\mathcal{L}_{\Dom,V}(E)$. \\
$V(p)$ & Program or algorithm evaluator & Route-specific evaluator for algorithmic discovery; higher values mean better candidate programs. \\
$\operatorname{Check}(\pi)$ & Proof checker & A strict verifier that accepts or rejects a proof certificate $\pi$. \\
$z=(h,c,e,r,p)$ & AI Scientist artifact tuple & Idea, code, experimental output, analysis, and manuscript in an end-to-end pipeline. \\
$R_{\mathrm{CSC}}(A')$ & Typed-artifact residual & A complementary measure of whether a new artifact exceeds what is transported from a prior schema. \\
$A'$, $I'_{t+1}$, $\rho_{A'}$ & Typed-artifact terms & New artifact, new interpretation, and schema-transport map in the CategoryScienceClaw comparison. \\
$\alpha,\beta,\gamma,\delta,\eta$ & Positive-value weights & Coefficients for prediction, causality, explanation, novelty, information gain, or related benefits. \\
$\rho,\lambda,\kappa,\mu,\nu$ & Penalty or auxiliary weights & Coefficients for novelty, cost, risk, complexity, and the overclaim penalty in the objective functions. \\
\end{longtable}
}

\section{A Unified Abstraction: Scientific Exploration as Policy Optimization}

Let the real world contain an unknown mechanism
\begin{equation}
  \omega^\ast \in \Omega,
\end{equation}
where $\Omega$ denotes the space of possible worlds, mechanisms, causal structures, physical laws, biological processes, or material stability rules. Scientists cannot directly observe $\omega^\ast$. They obtain data through literature, observation, experiment, simulation, theorem proving, and databases:
\begin{equation}
  D_t=\{(a_i,y_i,c_i)\}_{i=1}^{t}.
\end{equation}
Here $a_i$ denotes an experimental action, computational action, query, proof attempt, simulation condition, or intervention; $y_i$ denotes the observed outcome; and $c_i$ denotes contextual conditions such as cell type, temperature, solvent, genetic background, material composition, or benchmark setting.

Thus, throughout the paper, $D_t$ denotes the system's accumulated evidential state rather than a single dataset. The triple $(a_i,y_i,c_i)$ keeps three things separate: what the system did, what it observed, and where that observation is valid.

Scientific inquiry can be written as a policy:
\begin{equation}
  \pi:S_t\rightarrow a_{t+1},
\end{equation}
with state
\begin{equation}
  S_t=(D_t,\Know,H_t,M_t,B_t).
\end{equation}
$\Know$ is the prior knowledge base, $H_t$ is the set of hypotheses, $M_t$ is the current model, and $B_t(\omega)=P(\omega\mid D_t,\Know)$ is the current belief distribution over possible mechanisms. An idealized AI science system may be written as:
\begin{equation}
\begin{aligned}
\pi^\ast
&=\argmax_{\pi}\E\left[\sum_t \gamma^t\Phi(a_t,S_t)\right],\\
\Phi(a_t,S_t)
&=\alpha\Delta\mathrm{Pred}
+\beta\Delta\mathrm{Causal}
+\eta\Delta\mathrm{Explanation}
+\rho\Delta\mathrm{Novelty}\\
&\quad+\zeta\Delta\mathrm{Info}\\
&\quad-\lambda \Cost(a_t)-\kappa \Risk(a_t).
\end{aligned}
\end{equation}
The system should maximize predictive improvement, causal identification, explanatory power, novelty, and information gain under constraints of cost, safety, and feasibility.

In this objective, $\pi$ is not a language model prompt; it is the whole inquiry policy that chooses the next test. The function $\Phi$ is the local scientific value of a test, and the coefficients $\alpha,\beta,\eta,\rho,\zeta,\lambda,\kappa$ indicate how much the system values prediction, causality, explanation, novelty, information, cost, and risk. The explicit functions $\Cost$ and $\Risk$ avoid overloading $C$, which is reserved for scientific claims.

This expression is a schematic policy objective rather than a fully specified MDP or POMDP. A concrete transition kernel $P(S_{t+1}\mid S_t,a_t)$, observation model, and belief-update rule must be supplied by the domain. The schematic objective does not imply that every AI system must become a complete scientist. It allows us to identify which part of the scientific objective each route optimizes. Specialized models primarily optimize predictive improvement. LLM assistants expand the hypothesis space. Multi-agent systems optimize generation and critique. AI Scientist systems optimize research-artifact production. Formal systems optimize verifiable objects. Self-driving laboratories optimize information gain through real-world interventions.

\section{Scientific Semantics: From Text Continuation to World Constraint}

A scientific hypothesis cannot be reduced to a natural-language sentence. Consider:
\[
h=\text{``gene }g\text{ activates pathway }p\text{.''}
\]
At the linguistic level, this is just a sentence. At the scientific level, it must constrain which worlds, interventions, and observations are possible. Its model-theoretic meaning can be written as
\begin{equation}
  \denote{h}=\{\omega\in\Omega:\omega\models h\}.
\end{equation}
Scientific meaning, however, also requires testable consequences:
\begin{equation}
  \Sem_{\Dom}(h)=\Conseq_{\Dom}(h)
  =
  \{\chi_h(a,x):a\in\mathcal{A}_{\Dom},x\in\mathcal{X}_{\Dom}\},
  \qquad
  \chi_h(a,x)\in\mathcal{Q}_{\Dom}.
\end{equation}
In words, $\Conseq_{\Dom}(h)$ is the profile of domain-appropriate consequences that hypothesis $h$ implies under possible actions, queries, observations, or proof attempts. The space $\mathcal{Q}_{\Dom}$ may contain probability distributions, deterministic functions, constraint sets, proof obligations, counterexample conditions, diagnostic features, measurement signatures, classification rules, or retrodictive traces. In interventional experimental science, an important special case is:
\begin{equation}
  \chi_h(a,x)=P_h(Y\mid \doop(a),x).
\end{equation}
The notation $P_h(Y\mid \doop(a),x)$ follows the interventionist convention that the consequence is derived under the assumption that $h$ is true and that the action $a$ is performed rather than merely observed \citep{Pearl2009Causality}. The context variable is written as $x\in\mathcal{X}_{\Dom}$ here so that the symbol $C$ remains reserved for scientific claims.
For example, the hypothesis $h:g\rightarrow p$ should be expandable into:
\begin{equation}
\E[Y_p\mid \doop(g\uparrow),c]
-
\E[Y_p\mid \doop(g\downarrow),c]
>\delta.
\end{equation}
The meaning of a scientific hypothesis is therefore not whether a language model can explain it fluently, but whether the hypothesis can be translated into a structure that constrains observable, formal, diagnostic, measurement, or retrodictive consequences.

This paper defines the meaning of a scientific hypothesis as:
\begin{equation}
\boxed{
\mathrm{ScientificMeaning}_{\Dom}(h)=
\left(
\denote{h},
\Conseq_{\Dom}(h),
\mathcal{T}_{\Dom}(h),
\mathcal{V}_{\Dom}(h),
\mathcal{B}_{\Dom}(h)
\right)
}
\end{equation}
where $\mathcal{T}_{\Dom}(h)$ is the set of tests, observations, simulations, benchmarks, measurements, or proof obligations; $\mathcal{V}_{\Dom}(h)$ is the rule for verification, support, failure, or refutation; and $\mathcal{B}_{\Dom}(h)$ is the claim boundary: the range of assertions about $h$ that the current evidence permits. Without $\mathcal{B}_{\Dom}(h)$, scientific semantics remains incomplete. A hypothesis may have clear testable consequences while still leaving open whether the evidence licenses the language of possibility, preliminary support, mechanistic establishment, general law, or application-ready discovery.

The boxed tuple therefore states that a hypothesis has scientific meaning only when five components are specified: the worlds it allows, the consequences it implies, the tests it induces, the rule that evaluates those tests, and the boundary on what may be claimed after evaluation.

Ordinary LLMs mainly learn $P(\mathrm{text}_{t+1}\mid \mathrm{text}_{\leq t})$. Scientific semantics requires a family of consequences $P(Y\mid \doop(a),h,x)$ across actions and contexts. Evidential calibration semantics further requires a boundary on what can be asserted after validation. The first is linguistic continuation; the second is constraint on the world; the third is permission to make a bounded scientific claim.

\section{Evidential Calibration Semantics}

The preceding section treats the hypothesis $h$ as the central semantic object. Scientific papers, however, do not end with hypotheses. They end with claims. A hypothesis may be proposed, tested, supported, weakened, or rejected, but the manuscript must decide what it is permitted to say. This makes claim calibration a distinct epistemic operation rather than a stylistic afterthought.

The deeper point is that a scientific claim is not an ordinary sentence. It is an assertion act that requires warrant. Evidence does not merely increase confidence; it licenses certain forms of scientific speech and withholds others. A proof certificate, an in vitro assay, a simulation, a benchmark improvement, and an independent replication differ not only in strength, but also in the kinds of claims they authorize. Calibration semantics is therefore a theory of warranted assertability: under a given domain context, what assertion right has the system acquired?

The intended contribution is not to rename philosophical warrant, Toulmin-style warrants, or evidence-grading frameworks \citep{Toulmin1958Uses,Guyatt2008GRADE}. It is to make warrant operational for AI-assisted research systems whose outputs can shift between hypotheses, artifacts, figures, titles, abstracts, press releases, and user-facing claims. The additional object is an output contract: given a raw or implied claim, an evidence body, an evaluator, and a domain context, the system should expose the licensed frontier and the residual epistemic debt rather than merely produce plausible scientific prose.

For AI-system analysis, this section defines an output-contract layer for scientific agents. A scientific agent may generate many hypotheses, rank them, run code, search literature, call a simulator, submit jobs to a robotic laboratory, or write a manuscript. These capabilities still leave one question unresolved: which claim is licensed once evaluation is complete? Evidential calibration semantics answers that question by separating three objects that are often conflated in AI systems: the hypothesis being considered, the evidence that has been collected, and the claim that the system makes to the user.

This distinction matters because AI systems tend to produce fluent assertions even when their evidence is heterogeneous. A benchmark improvement, a proof certificate, a simulation, a single wet-lab assay, and a multi-site replication are all forms of evidence, but they do not license the same language. A system that treats all successful validation as equivalent may report a claim that is too strong. The purpose of the calibration layer is to make this failure mode visible and correctable. In this sense, $\Cal_{\Dom}$ is not a module for making sentences sound more cautious. It is the mechanism that determines whether an AI system has the epistemic right to assert a scientific sentence in domain context $\Dom$.

This section has three layers. First, it separates hypotheses, claims, literal wording, implied framing, and evidence. Second, it formalizes evidence-licensed assertion through a domain-indexed license relation. Third, it defines calibration as a projection from raw claims to the maximal evidence-licensed frontier. The notation is deliberately repeated near each definition because the section is the theoretical center of the paper.

\subsection{Claims Are Not Hypotheses}

Let a scientific claim be written as:
\begin{equation}
C=\Claim(h,s,b,m),
\end{equation}
where $h$ is the hypothesis, $s$ is claim strength, $b$ is the boundary of applicability, and $m$ is the modal status of the assertion: possible, supported, demonstrated, explained, discovered, or translated. The same hypothesis can license very different claims:
\begin{itemize}
  \item a weak claim: an AI system proposed an experimentally testable candidate mechanism;
  \item a moderate claim: the candidate mechanism received in vitro support under specified cell-line and assay conditions;
  \item a strong claim: the mechanism establishes a general disease principle;
  \item an overstrong claim: the system discovered a clinically validated therapy.
\end{itemize}
These are not equivalent formulations. They differ in the evidence required for their assertion.

In an AI system, the claim object $C$ should be treated as a first-class output, not as a prose afterthought. The system may internally manipulate embeddings, programs, proofs, experimental protocols, or molecular structures, but the external scientific product is usually a claim expressed in language. The representation $\Claim(h,s,b,m)$ forces that claim to expose four components: what hypothesis it refers to, how strong it is, where it applies, and what modal verb or epistemic status it uses.

Calibration is therefore not merely weakening. It is the joint adjustment of strength, scope, modality, and mechanistic content. In structured form:
\begin{equation}
\Cal_{\Dom}(\Claim(h,s,b,m),E,V)
=\Claim(h',s',b',m'),
\end{equation}
where the formal requirement is that the whole calibrated claim satisfies
\[
\Claim(h',s',b',m')\preceq_{\Dom}\Claim(h,s,b,m).
\]
This claim order may be induced by weaker strength, narrower boundary, weaker modal status, or a less mechanistic hypothesis object, but the paper does not require every component to share a single primitive order. For example, a raw claim about a causal mechanism may become a calibrated claim about a predictive association; a raw translational claim may become a claim about selected cell-line settings; a raw discovery claim may become a claim about generated candidates consistent with current evidence.

Calibration should also apply to implied claims, not only literal sentences. Let $C_{\mathrm{literal}}$ denote the exact sentence in a paper, report, title, or system output, and let $C_{\mathrm{implied}}$ denote the stronger or broader claim induced by the title, abstract, figure design, framing, press release, or downstream communication. It is possible that
\begin{equation}
C_{\mathrm{literal}}\in\mathcal{L}_{\Dom,V}(E)
\quad\text{but}\quad
C_{\mathrm{implied}}\notin\mathcal{L}_{\Dom,V}(E).
\end{equation}
This is why calibration is not only a sentence-level editing rule. It is a publication-level and system-output-level discipline.

\subsection{License Relations and Evidence Preorders}

Let the domain-specific claim space be a partially ordered set:
\begin{equation}
(\mathcal{C}_{\Dom},\preceq_{\Dom}),
\end{equation}
where $C_1\preceq_{\Dom} C_2$ means that $C_1$ is no stronger than $C_2$ in domain context $\Dom$. Strength is not only a matter of adjective choice. It depends on the claim's boundary, modal status, intended population, mechanism, intervention, benchmark, material class, or formalization.

Let the evidence space be:
\begin{equation}
\mathcal{E}_{\Dom},
\end{equation}
the set of admissible evidence bodies in domain $\Dom$. Evidence is not assumed to have a single global strength order. Instead, evidence comparisons are claim-relative. For a comparison family $\mathcal{C}_0\subseteq\mathcal{C}_{\Dom}$, write
\[
E_1\sqsubseteq_{\Dom,V,\mathcal{C}_0}E_2
\quad\Longleftrightarrow\quad
\forall C\in\mathcal{C}_0,\ 
\Sup_{\Dom,V}(E_2,C)\succeq_{\mathcal{W},\Dom}\Sup_{\Dom,V}(E_1,C).
\]
This says that $E_2$ is no weaker than $E_1$ for the claims being compared. Formally, $\sqsubseteq_{\Dom,V,\mathcal{C}_0}$ is a preorder on evidence bodies, not necessarily a partial order: two distinct evidence packages can induce the same support profile over $\mathcal{C}_0$. If a partial order is needed, one can quotient by the equivalence relation $E_1\sim E_2$ when $E_1\sqsubseteq E_2$ and $E_2\sqsubseteq E_1$. One independent replication may dominate many non-independent benchmark runs for a biological mechanism claim; a proof certificate may dominate extensive informal argument for a formal theorem; neither necessarily dominates the other for all possible claims.

The support and requirement profiles live in a warrant-profile space:
\begin{equation}
(\mathcal{W}_{\Dom},\succeq_{\mathcal{W},\Dom}).
\end{equation}
The requirement map and support map are typed as
\begin{equation}
\Req_{\Dom}:\mathcal{C}_{\Dom}\rightarrow\mathcal{W}_{\Dom},
\qquad
\Sup_{\Dom,V}:\mathcal{E}_{\Dom}\times\mathcal{C}_{\Dom}\rightarrow\mathcal{W}_{\Dom}.
\end{equation}
Here $V$ appears in the support function because evaluator provenance, independence, reliability, and grounding affect what the evidence supports. Throughout this manuscript, the evaluator is kept explicit in the license relation unless a domain-specific implementation encodes evaluator provenance directly inside the evidence object.

The central relation is:
\begin{equation}
E\Vdash_{\Dom,V} C,
\end{equation}
This relation means that, under domain context $\Dom$ and evaluator $V$, evidence $E$ licenses claim $C$. It is a license relation, not a truth relation. A claim may be true but not yet licensed by the available evidence; a claim may be licensed by current evidence and later withdrawn when stronger evidence appears. Scientific practice does not directly possess truth. It acquires defensible assertion rights under fallible but auditable evidential norms.

\subsection{The Claim-Evidence Gap and Epistemic Debt}

Given a body of evidence $E$, an evaluator $V$, and a claim $C$, first define the support-coverage condition as:
\begin{equation}
\Sup_{\Dom,V}(E,C)\succeq_{\mathcal{W},\Dom} \Req_{\Dom}(C).
\end{equation}
Here $\Req_{\Dom}(C)$ is the evidential requirement profile imposed by the claim, $\Sup_{\Dom,V}(E,C)$ is the support profile that the available evidence and evaluator provide for that claim, and $\succeq_{\mathcal{W},\Dom}$ means that the support profile covers the requirement profile. The primary licensing relation is therefore not a universal scalar score, but a domain-specific comparison between evidential profiles:
\begin{equation}
E\Vdash_{\Dom,V} C
\quad\Longleftrightarrow\quad
\Sup_{\Dom,V}(E,C)\succeq_{\mathcal{W},\Dom} \Req_{\Dom}(C).
\end{equation}
If $\Sup_{\Dom,V}(E,C)\not\succeq_{\mathcal{W},\Dom}\Req_{\Dom}(C)$, the claim exceeds the evidence.

For engineering use, the profile comparison can be projected to a scalar diagnostic:
\begin{equation}
\Delta_{\Dom,V}(C,E)=\ell_{\Dom}(\Req_{\Dom}(C))-\ell_{\Dom}(\Sup_{\Dom,V}(E,C)),
\end{equation}
where $\ell_{\Dom}:\mathcal{W}_{\Dom}\rightarrow\mathbb{R}$ is an order-preserving scalarization used only for diagnostics:
\[
w_1\succeq_{\mathcal{W},\Dom}w_2
\Rightarrow
\ell_{\Dom}(w_1)\geq \ell_{\Dom}(w_2).
\]
Under such a projection, $\Delta_{\Dom,V}(C,E)>0$ means that the claim is stronger than the evidence; $\Delta_{\Dom,V}(C,E)\leq 0$ means that the chosen scoring map treats the evidence as sufficient for that level of assertion. The scalar gap is useful for audits and dashboards, but it should not be mistaken for a universal theory of evidence.

The formula has a right-to-left interpretation. First ask what evidence the claim requires, $\Req_{\Dom}(C)$. Then ask what the actual evidence and evaluator provide, $\Sup_{\Dom,V}(E,C)$. The gap $\Delta_{\Dom,V}(C,E)$ is the overclaiming pressure that the paper tries to make explicit after a domain has chosen how to score or order evidence.

This distinction is essential because validation does not have a single semantic value. An in vitro result, a benchmark improvement, a formal proof, a simulation, and an independent cohort replication all count as validation in some sense, but they license different claims. The central question becomes: what does this evidence permit the system to say?

For implementation, $\Delta_{\Dom,V}(C,E)$ can be interpreted as a diagnostic signal. If $\Delta_{\Dom,V}(C,E)$ is positive, the generated claim should be revised before being shown as a scientific conclusion. If $\Delta_{\Dom,V}(C,E)$ is near zero, the system is operating near the edge of what the evidence licenses and should report boundary conditions clearly. If $\Delta_{\Dom,V}(C,E)$ is negative, the evidence is at least sufficient for the current level of assertion, although it may still be insufficient for a stronger claim.

Overclaiming is not merely an error in wording. It is epistemic debt:
\begin{equation}
\Debt^\ell_{\Dom,V}(C,E)=\max\{0,\Delta_{\Dom,V}(C,E)\}.
\end{equation}
This is scalarized epistemic debt, because it depends on the diagnostic map $\ell_{\Dom}$. A profile-level debt object, $\Debt^{\mathrm{profile}}_{\Dom,V}(C,E)$, can also be defined by each domain as the unmet part of $\Req_{\Dom}(C)$ after comparison with $\Sup_{\Dom,V}(E,C)$, but there is no domain-independent subtraction operation on warrant profiles. A scientific system can pay this debt by collecting more evidence, such as a new experiment, replication, proof certificate, benchmark, or independent cohort. It can also cancel the debt by weakening the claim, narrowing its boundary, changing its modal status, or revising the hypothesis object. A system that does neither accumulates epistemic debt while presenting a stronger assertion than it has earned.

\subsection{Licensed Claim Sets and the Calibration Operator}

Define the set of claims licensed by evidence $E$ as:
\begin{equation}
\mathcal{L}_{\Dom,V}(E)=\{C\in\mathcal{C}_{\Dom}:E\Vdash_{\Dom,V} C\}.
\end{equation}
This set should be downward closed in the claim poset, provided both requirement monotonicity and claim-relative support monotonicity hold. Requirement monotonicity says:
\begin{equation}
C'\preceq_{\Dom} C
\quad\Rightarrow\quad
\Req_{\Dom}(C')\preceq_{\mathcal{W},\Dom}\Req_{\Dom}(C),
\end{equation}
where $\preceq_{\mathcal{W},\Dom}$ is the reverse of $\succeq_{\mathcal{W},\Dom}$ and means ``requires no more warrant than''. Claim-relative support monotonicity says:
\begin{equation}
C'\preceq_{\Dom} C
\quad\Rightarrow\quad
\Sup_{\Dom,V}(E,C')\succeq_{\mathcal{W},\Dom}\Sup_{\Dom,V}(E,C).
\end{equation}
This means that the same evidence is not less supportive of a weaker version of the same claim. Equivalently, a domain may take downward license monotonicity as a primitive norm:
\begin{equation}
C\in\mathcal{L}_{\Dom,V}(E),\quad C'\preceq_{\Dom} C
\quad\Rightarrow\quad
C'\in\mathcal{L}_{\Dom,V}(E).
\end{equation}
In words: if the evidence licenses a strong claim, it also licenses all weaker claims, assuming weaker claims require no stronger warrant than stronger claims and the same evidence is not less supportive of a weaker assertion. Thus $\mathcal{L}_{\Dom,V}(E)$ is a lower set of the domain-specific claim space.

To make calibration total, assume that each domain has a bottom claim $\bot_{\Dom}\in\mathcal{C}_{\Dom}$, such as ``no scientific claim is licensed beyond proposal.'' The bottom claim satisfies $\bot_{\Dom}\preceq_{\Dom}C$ for all $C\in\mathcal{C}_{\Dom}$ and is always licensed at the minimal proposal level, $\bot_{\Dom}\in\mathcal{L}_{\Dom,V}(E)$. This prevents the calibration set from being empty even when the raw claim is entirely unsupported.

Let the down-set of the raw claim be:
\begin{equation}
\downarrow C=\{C'\in\mathcal{C}_{\Dom}:C'\preceq_{\Dom} C\}.
\end{equation}
The task of claim calibration is to find the maximal licensed weakening frontier of the raw claim:
\begin{equation}
\Cal_{\Dom}(C,E,V)
=\Max_{\preceq_{\Dom}}\left(\downarrow C\cap\mathcal{L}_{\Dom,V}(E)\right).
\end{equation}
Thus $\Cal_{\Dom}$ is set-valued in general:
\[
\Cal_{\Dom}:\mathcal{C}_{\Dom}\times\mathcal{E}_{\Dom}\times\mathcal{V}_{\Dom}
\rightarrow \mathcal{P}(\mathcal{C}_{\Dom}).
\]
Because $\bot_{\Dom}\in\downarrow C\cap\mathcal{L}_{\Dom,V}(E)$, the intersection is non-empty. If it has a unique maximum, the frontier contains one calibrated claim. If there are several incomparable maximal claims, $\Cal_{\Dom}$ returns a frontier rather than a single sentence. A domain-specific selection rule $\sigma_{\Dom}$ may then choose a reportable claim,
\[
C^\ast=\sigma_{\Dom}(\Cal_{\Dom}(C,E,V)),
\]
according to domain convention, audience, and reporting purpose, but it should not step outside the frontier.

In practical reporting settings, the claim space is implemented as a finite or discretized reporting lattice, such as the L0--L6 ladder introduced below, so maximal licensed weakenings exist. More abstractly, the framework assumes that each feasible set considered by $\Cal_{\Dom}$ has at least one maximal element.

The important output is the licensed frontier, and then the selected claim $C^\ast$ if the system or author needs a single sentence. $C^\ast$ is allowed to be weaker than the original claim $C$; in fact, downgrading an overstrong claim is one function of $\Cal_{\Dom}$. But $\Cal_{\Dom}$ can also narrow the boundary, change the assertion modality, or revise the mechanistic object. Calibration is therefore a semantic transformation, not a conservative language filter.

\paragraph{Structural compression and minimal reconstruction.}
A less obvious form of calibration is not a downward weakening but an upward change of level. Several heterogeneous outputs may each support only a local claim, yet together suggest a common structure. In model selection, the minimum-description-length tradition treats learning as the extraction of regularity that supports a shorter description of data \citep{Rissanen1978ShortestDataDescription,GrunwaldRoos2019MDLRevisited}. In philosophy of science, unification accounts treat explanation as reducing heterogeneous phenomena to a smaller family of explanatory patterns \citep{Friedman1974ExplanationUnderstanding,Kitcher1981ExplanatoryUnification}. In mathematics, understanding often consists not only in checking isolated propositions, but in finding reusable methods, definitions, and ways of thinking that reconstruct many results at once \citep{Thurston1994ProofProgress,Avigad2006MathematicalMethodProof}. These traditions suggest a special claim-calibration problem: the strongest licensed contribution may be neither a narrower local claim nor a broad unqualified discovery claim, but a minimal structural reconstruction of what multiple local claims have in common. In this sense, structural compression is claim calibration across heterogeneous contents: it asks what common claim, if any, can be asserted over a family rather than over each member separately.

Formally, let $\mathcal{F}=\{C_1,\ldots,C_k\}\subseteq\mathcal{C}_{\Dom}$ be a comparison family of claims, and let $E_{\mathcal{F}}$ and $V_{\mathcal{F}}$ record the evidence and evaluator provenance for that family. A candidate compressed claim is a minimal structural reconstruction when it is a minimal upper claim for the family:
\begin{equation}
C_{\mathrm{str}}\in
\operatorname{Min}_{\preceq_{\Dom}}
\{C\in\mathcal{C}_{\Dom}:\forall i,\ C_i\preceq_{\Dom} C\}.
\end{equation}
It becomes assertable only through the ordinary calibration operator:
\begin{equation}
C_{\mathrm{str}}^\ast\in
\Cal_{\Dom}(C_{\mathrm{str}},E_{\mathcal{F}},V_{\mathcal{F}}).
\end{equation}
Here ``upper'' names the level of reconstruction relative to the comparison family, not an automatic increase in evidential authority: the reconstructed claim may speak at a broader structural level only to the extent that $E_{\mathcal{F}}$ and $V_{\mathcal{F}}$ license the shared structure and leave route-specific residuals explicit. Non-amplification applies after $C_{\mathrm{str}}$ has been posed as a new family-level raw claim; it does not permit $\Cal_{\Dom}(C_i,E_i,V_i)$ to strengthen an individual local claim without additional family-level evidence. This is an upward or structural use of calibration. Compression may organize heterogeneous licensed claims, but any additional generality, causal interpretation, cross-context scope, or discovery language introduced by the reconstruction must itself be licensed. A shorter common structure is therefore not automatically a truer or stronger scientific claim; it is a candidate structural claim whose assertion rights depend on the family-level evidence and the residual differences that the reconstruction leaves outside its scope.

This is why $\Cal_{\Dom}$ is separated from $V$. The evaluator $V$ answers whether a test was passed, a benchmark improved, a proof checked, or an assay produced the expected response. The calibration operator $\Cal_{\Dom}$ answers a different question: given that result, what assertion right has the system acquired? A single positive test may justify ``is consistent with'', ``supports'', or ``demonstrates under these conditions'', but not necessarily ``discovers'', ``establishes'', or ``translates to clinical use''.

\paragraph{Minimal desiderata for calibration.}
A calibration operator should satisfy at least four conditions. First, it should be sound:
\begin{equation}
\Cal_{\Dom}(C,E,V)\subseteq \mathcal{L}_{\Dom,V}(E),
\end{equation}
Second, it should be non-amplifying:
\begin{equation}
\forall C'\in\Cal_{\Dom}(C,E,V),\quad C'\preceq_{\Dom} C,
\end{equation}
so calibration may weaken an overstrong claim but should not strengthen it without additional evidence. Third, it should be monotone with respect to evidential support:
\begin{equation}
\left[
\forall C'\preceq_{\Dom} C,\
\Sup_{\Dom,V}(E',C')\succeq_{\mathcal{W},\Dom}\Sup_{\Dom,V}(E,C')
\right]
\Rightarrow
C_E^\ast\preceq_{\Dom} C_{E'}^\ast,
\end{equation}
where $C_E^\ast=\sigma_{\Dom}(\Cal_{\Dom}(C,E,V))$ and $C_{E'}^\ast=\sigma_{\Dom}(\Cal_{\Dom}(C,E',V))$ are selected maximal elements under a fixed domain selection rule. That is, if a later evidence body provides uniformly stronger support for every weakening of the same raw claim, the selected calibrated claim should not become weaker. In the set-valued version, the downward closure of the licensed frontier should not shrink. Stronger evidence may license the same or a stronger calibrated claim, but should not force a downgrade unless the domain context or evaluator changes. Fourth, it should preserve boundaries: the calibrated claim must state the domain, assay, benchmark, population, material class, or formalization under which the evidence was obtained. These conditions make $\Cal_{\Dom}$ an assertion-right layer rather than a stylistic rewriting module.

When the maximal frontier is not unique, the monotonicity statement assumes a fixed selection rule compatible with $\Strength$. Equivalently, the set-valued interpretation is that stronger evidence should not shrink the downward closure of the licensed frontier for the same raw claim and domain context.

The resulting theory has four nested levels. The semantic level specifies
\[
\mathrm{ScientificMeaning}_{\Dom}(h)=
(\denote{h},\Conseq_{\Dom}(h),\mathcal{T}_{\Dom}(h),\mathcal{V}_{\Dom}(h),\mathcal{B}_{\Dom}(h)).
\]
The license level asks whether $E\Vdash_{\Dom,V} C$. The calibration level computes
\[
\Cal_{\Dom}(C,E,V)=\Max_{\preceq_{\Dom}}(\downarrow C\cap\mathcal{L}_{\Dom,V}(E)).
\]
The dynamical level, introduced later, asks whether generation and consequence-derivation capacity grow faster than adjudication and calibration capacity. This four-layer structure is the bridge from semantic theory to AI-system design.

\subsection{Dimensions of Evidential Support}

The support profile $\Sup_{\Dom,V}(E,C)$ is not one-dimensional. It may be decomposed as:
\begin{equation}
\Sup_{\Dom,V}(E,C)=F_{\Dom,V}(d,i,r,e,c_{\mathrm{id}},\xi,n),
\end{equation}
where $d$ is directness of the test, $i$ is independence from the generating system, $r$ is reproducibility, $e$ is effect size, $c_{\mathrm{id}}$ is causal identification, $\xi$ is external validity, and $n$ captures negative evidence, failed cases, or counterexamples. This decomposition turns claim calibration from a vague philosophical caution into an auditable checklist.

Thus $\Sup_{\Dom,V}(E,C)$ is not just a confidence score. A claim can have a large effect size but weak external validity, or strong computational support but little independence from the generating system. In this paper, $\Sup_{\Dom,V}(E,C)$ is treated as a domain- and evaluator-specific warrant functional rather than a universal scalar. Different domains may instantiate $F_{\Dom,V}$ as a checklist, ordinal rubric, Bayesian score, proof-verification predicate, benchmark protocol, or expert-audited evidence profile. The calibration operator should see these differences before it licenses a claim.

The ladder is not intended as a universal hierarchy of all sciences. It is a reporting template: each domain must instantiate what counts as support, replication, causality, generalization, and application. A formal proof, a wet-lab assay, a benchmark result, and a materials synthesis campaign occupy different evidential geometries even when they are all called validation.

\begin{table}[htbp]
\centering
\small
\caption{Domain-parametric claim ladder for AI-driven scientific assertions}
\begin{tabularx}{\textwidth}{@{}p{1.1cm}p{3cm}Y Y@{}}
\toprule
Level & Claim type & Licensed wording & Required evidence \\
\midrule
L0 & Hypothesis generation & AI proposed a possible mechanism or candidate & Literature consistency and conceptual plausibility \\
L1 & Computational support & The mechanism is supported by a model or simulation & Computational validation, baselines, ablations, leakage checks \\
L2 & Internal experimental support & The mechanism is supported in the present experimental system & Controlled experiment, statistics, assay controls \\
L3 & External replication support & The mechanism is supported by independent data or experiment & Independent cohort, independent laboratory, or independent dataset \\
L4 & Causal mechanism support & The mechanism is stable under intervention & Perturbation, causal identification, negative controls \\
L5 & Generalizable scientific knowledge & The regularity holds across conditions within stated boundaries & Multi-context replication and explicit boundary conditions \\
L6 & Translational or application claim & The finding can support a drug, material, or engineering use & Safety, efficacy, deployment constraints, external validation \\
\bottomrule
\end{tabularx}
\end{table}

Much of the evidence currently reported in AI science systems lies around L1--L3. Public-facing language, however, can drift toward L4--L6. The role of evidential calibration semantics is to prevent this drift by forcing the final claim to remain inside $\mathcal{L}_{\Dom,V}(E)$.

The ladder is domain-parametric rather than universal. In mathematics, a proof accepted by a trusted formal checker may license theorem-level claims without experimental replication. In biology, in vitro support does not license a translational or clinical claim. In machine learning, a benchmark improvement may license a constrained performance claim, but not necessarily a mechanistic discovery claim. In materials science, synthesis of a target compound is not automatically equivalent to discovery of a new material unless phase identification, novelty, and reproducibility have also been established. The same ladder therefore functions as a template whose evidential thresholds must be instantiated by domain-specific norms, similar in spirit to evidence-grading frameworks in medicine \citep{Guyatt2008GRADE}.

\subsection{Worked Example: Co-Scientist and AML Drug Repurposing}

Consider an overstrong reading: Co-Scientist discovered new AML treatments. The evidence reported for Co-Scientist is more specific: the system generated drug-repurposing candidates and synergistic combination hypotheses for acute myeloid leukemia, with selected candidates validated in vitro \citep{Gottweis2026CoScientist}. A calibrated claim is therefore:
\begin{quote}
Co-Scientist generated drug-repurposing candidates for AML, and selected candidates showed in vitro anti-tumor activity in specific AML cell-line settings under expert-guided validation.
\end{quote}
This claim is stronger than mere hypothesis generation, because it includes reported in vitro support beyond model output. It is weaker than saying that Co-Scientist discovered a clinically validated AML therapy. The difference is not rhetorical. It is the difference between within-study laboratory validation under specified assay conditions and a translational or clinically validated therapeutic claim.

\subsection{Worked Example: AI Scientist and Research Artifacts}

Consider a strong raw claim: The AI Scientist autonomously performs scientific discovery. The evidence reported for AI Scientist systems is more constrained: the system can generate machine-learning ideas, write code, execute experiments, produce figures, draft manuscripts, and obtain workshop-level review outcomes under a restricted computational research setting \citep{Lu2026AIScientist}. A calibrated claim is therefore:
\begin{quote}
The AI Scientist automates substantial portions of machine-learning research artifact generation and can produce workshop-level manuscripts under constrained benchmark and review settings.
\end{quote}
This calibrated claim recognizes the system's contribution without upgrading artifact production into general autonomous science. In this case, $\Cal_{\Dom}$ does not deny the system's automation capability; it prevents workflow completion and workshop-level review from being redescribed as general scientific autonomy. The claim may be strong at the level of workflow automation, but it remains weaker than saying that the system has established a reliable general mechanism for producing novel scientific knowledge across domains.

\subsection{Worked Example: A-Lab and Materials Claims}

Consider a commonly circulated strong claim: A-Lab discovered 43 new materials. After the correction and subsequent discussion, the more careful evidential reading is narrower: the autonomous laboratory realized a set of target compounds under specified solid-state synthesis and characterization criteria, while claims of novelty require phase identification, database deduplication, and independent review \citep{Szymanski2023ALab,Szymanski2026ALabCorrection,Chawla2026ALabCorrection}. A calibrated claim is therefore:
\begin{quote}
A-Lab demonstrated an automated workflow for proposing, synthesizing, and characterizing inorganic target compounds over a short campaign, while the strength of new-material discovery claims depends on corrected phase assignment, novelty checks, and independent materials review.
\end{quote}
This example shows why world contact is not itself sufficient for claim licensing. A robotic laboratory can produce real experimental evidence, but the final scientific claim still depends on how the evidence is classified, compared with prior databases, and bounded by independent expertise.

\subsection{Relation to Typed Artifact Discovery}

Frameworks such as CategoryScienceClaw, as proposed in the arXiv preprint on self-revising discovery systems by \citet{WangBuehler2026SelfRevising}, solve a complementary problem: they formalize typed artifacts, provenance, verifiers, and regime transitions in agentic scientific discovery systems. In such a typed-artifact view, discovery is modeled as a regime transition in which scientific artifacts, verifiers, provenance, and schema revisions are preserved and audited. A schematic residual can be written as:
\begin{equation}
R_{\mathrm{CSC}}(A')=
I'_{t+1}(A')\setminus \operatorname{im}(\rho_{A'}),
\end{equation}
which measures whether a new artifact exceeds what can be transported from the previous schema. The present framework instead asks whether a scientific claim exceeds what the evidence licenses:
\begin{equation}
\Delta_{\Dom,V}(C,E)=\ell_{\Dom}(\Req_{\Dom}(C))-\ell_{\Dom}(\Sup_{\Dom,V}(E,C)).
\end{equation}
The former is an artifact-regime residual; the latter is a claim-evidence gap. A new artifact is not yet a licensed claim. Typed provenance can show where a claim came from, but not by itself how strong the claim is allowed to be. Typed-artifact systems address how scientific workflows can be typed, preserved, and audited. Evidential semantics addresses how scientific claims are permitted, restricted, or downgraded by evidence.

\section{Six Routes and Their Epistemological Structure}

\subsection{Specialized Scientific Foundation Models}

Specialized scientific foundation models are not autonomous scientists. They are strong predictors for particular scientific problems. AlphaFold 3 targets biomolecular complex structure prediction. The AlphaFold Protein Structure Database is an open database of predicted protein structures and covers more than 200 million predicted entries \citep{Varadi2024AlphaFoldDB,AlphaFoldDB2026}. In materials search, \citet{Merchant2023GNoME} used GNoME to computationally identify 2.2 million structures stable with respect to the Materials Project, of which 381,000 lie on the updated convex hull.

The basic form is supervised learning or generative modeling:
\begin{equation}
\theta^\ast=\argmin_{\theta}\E_{(x,y)\sim \mathcal{D}_{\mathrm{train}}}[\ell(f_{\theta}(x),y)],
\end{equation}
or probabilistically $p_{\theta}(y\mid x)$. In materials discovery one may write:
\begin{equation}
x^\ast=\argmax_x P_{\theta}(\mathrm{stable}\mid x).
\end{equation}
The implicit epistemology is:
\[
\boxed{\text{to know} \approx \text{to predict structure, energy, or property accurately}.}
\]
The advantage is high-throughput screening: $|X|\gg |X_{\mathrm{candidate}}|$. The risk is:
\begin{equation}
p_{\theta}(y\mid x)\neq P_{\mathrm{world}}(y\mid \doop(x)).
\end{equation}
A prediction is not an explanation, and a candidate is not a validation. These models are best understood as search-space compressors, not as final scientific justifiers.

\subsection{Human-Led LLM Research Assistants}

An LLM research assistant may be written as:
\begin{equation}
h\sim P_{\phi}(h\mid \Know,q,D_t),
\end{equation}
where $\Know$ is the literature and knowledge base, $q$ is the research question, $D_t$ is available data accumulated up to time $t$, and $h$ is a hypothesis, explanation, experimental suggestion, or proof sketch. A rough optimization form is:
\begin{equation}
h^\ast=\argmax_h
\left[
\log P_{\phi}(h\mid \Know,q,D_t)
+\lambda S_{\mathrm{coherence}}(h)
+\mu S_{\mathrm{novelty}}(h)
\right].
\end{equation}
The GPT-5 science acceleration report presents cases in which frontier LLMs helped experts connect literature, draft proofs, explore computations, and generate testable hypotheses. The same report stresses that the examples are curated, not a systematic sample, and that models may still hallucinate citations, mechanisms, or proofs \citep{Bubeck2025GPT5Science,OpenAI2025ScienceGPT5}.

The underlying view of science is:
\[
\boxed{\text{scientific discovery} \approx \text{recombining high-value explanations from prior knowledge}.}
\]
The strength of this route is $\Know\rightarrow h$: the generation of hypotheses from a knowledge base. Its weakness is $h\rightarrow \Conseq_{\Dom}(h)$. LLM semantics is closer to $\Sem_{\mathrm{LLM}}(h)=\mathrm{embedding}(h,\Know)$, whereas scientific semantics requires a domain-appropriate consequence profile such as $\Conseq_{\Dom}(h)$, with $P_h(Y\mid \doop(a),c)$ as an interventional special case. LLM assistants should therefore be treated as human-in-the-loop exploration expanders, not as independent bearers of truth.

\subsection{Multi-Agent Co-Scientist Systems}

Multi-agent co-scientist systems treat science as a cycle of hypothesis generation, critique, rewriting, ranking, and experimental planning. Let $H_t=\{h_1,\ldots,h_n\}$ be the current hypothesis population. Define a generator $G_{\mathrm{gen}}$, critic $C_{\mathrm{crit}}$, reviser $R_{\mathrm{rev}}$, and selector $S_{\mathrm{sel}}$:
\begin{equation}
H_{t+1}=S_{\mathrm{sel}}\circ R_{\mathrm{rev}}\circ C_{\mathrm{crit}}\circ G_{\mathrm{gen}}(H_t,\Know,q,D_t).
\end{equation}
A hypothesis score can be written as:
\begin{equation}
s(h)=
\lambda_1N(h)+\lambda_2P(h\mid \Know,D_t)+\lambda_3T(h)+
\lambda_4F(h)+\lambda_5I(h;Y)-\lambda_6C_{\mathrm{test}}(h),
\end{equation}
where $N$ is novelty, $T$ is testability, $F$ is feasibility, and $I(h;Y)$ is expected information gain.

\citet{Gottweis2026CoScientist} describe Google Co-Scientist as a Gemini-based multi-agent system that continuously generates, critiques, refines, and evolves hypotheses using tournament evolution. Google Co-Scientist is distinct from Boiko et al.'s earlier Coscientist chemistry system, which is discussed below under self-driving laboratories. They evaluate Google Co-Scientist in biomedical applications including drug repurposing, target discovery, and antimicrobial resistance explanation, with AML drug-repurposing candidates and synergistic combinations tested in vitro. FutureHouse Robin integrates literature-search and data-analysis agents for experimental biology, including hypothesis generation, experimental suggestions, interpretation of results, and hypothesis updates \citep{Ghareeb2026Robin}.

This route combines abductive inference, Popperian falsifiability, and social epistemology. Its importance is not that it merely simulates debate inside a scientific community, but that it pushes multi-agent hypothesis generation toward experimental actionability. It can generate, criticize, rank, and translate hypotheses into tests. The more precise epistemological question is therefore not whether there is any validation, but which level of claim the validation supports.

If evidence shows that selected AI-generated candidates are effective under specified in vitro conditions, the licensed claim is that the system generated candidates that received in vitro support in expert-guided settings. This does not automatically license stronger claims, such as autonomous scientific discovery in general or clinically translatable therapy. Co-scientist systems therefore expose a calibration problem: experimental actionability can be mistaken for a broader scientific discovery unless $\Cal_{\Dom}$ explicitly maps evidence to a bounded claim.

\subsection{End-to-End AI Scientist Pipelines}

The AI Scientist route turns a research trajectory into:
\begin{equation}
z=(h,c,e,r,p),
\end{equation}
where $h$ is an idea, $c$ is code, $e$ is experimental output, $r$ is analysis, and $p$ is the manuscript. The objective can be written as:
\begin{equation}
J(z)=
\lambda_1S_{\mathrm{benchmark}}(e)
+\lambda_2S_{\mathrm{novelty}}(h)
+\lambda_3S_{\mathrm{paper}}(p)
+\lambda_4S_{\mathrm{review}}(p)
-\lambda_5C(z).
\end{equation}
\citet{Lu2026AIScientist} describe The AI Scientist as a system that creates research ideas, writes code, runs experiments, plots and analyzes data, writes manuscripts, and performs its own peer review. One generated manuscript exceeded the average human acceptance threshold in the first round of review for an ICLR workshop, but the authors explicitly note that the workshop had a lower bar than a main conference track.

The implicit epistemology is:
\[
\boxed{\text{scientific research} \approx \text{an executable workflow that can be written and reviewed}.}
\]
The advantage is speed and process automation, especially in machine learning, simulation, and low-cost computational experiments. The risk is Goodhart's law:
\begin{equation}
\max S_{\mathrm{paper}}\not\Rightarrow \max \mathrm{Truth}.
\end{equation}
If the system optimizes paper form, small benchmark improvements, and automatic review scores, it may generate research that is formally complete but epistemically weak.

\subsection{Algorithmic and Mathematical Discovery Agents}

For AlphaEvolve-like systems, the object is not a textual hypothesis but a program, algorithm, or proof. Let $p\in\mathcal{P}$ and $V:\mathcal{P}\rightarrow\mathbb{R}$. The target is:
\begin{equation}
p^\ast=\argmax_{p\in\mathcal{P}} V(p).
\end{equation}
For proof tasks, the target is a certificate $\pi$ such that:
\begin{equation}
\exists\pi:\operatorname{Verifier}(\pi,T)=1.
\end{equation}
AlphaEvolve combines Gemini-based generation, automated evaluator feedback, and evolutionary search for selected machine-gradable tasks. Its white paper describes applications to data-center scheduling, circuit simplification for hardware accelerators, LLM training efficiency, matrix multiplication, and problems in mathematics and computer science \citep{Novikov2025AlphaEvolve}. AlphaProof and AlphaGeometry 2 solved four of six IMO 2024 problems, scoring 28 out of 42 points, in the silver-medal range; the reported problem-specific reasoning substantially exceeded human contest time limits \citep{Hubert2026AlphaProof}.

The scientific view is:
\[
\boxed{\text{to know} \approx \text{to construct an object accepted by a verifier}.}
\]
If the evaluator is reliable, $V(p)=1$ or $\operatorname{Check}(\pi,T)=1$ provides a strong epistemic warrant. The limitation is equally clear:
\[
\mathrm{science\ domain}\subseteq \mathrm{domains\ with\ evaluators}.
\]
Algorithms, proofs, code, compiler optimization, and benchmarks suit this route. Complex biology, clinical medicine, ecosystems, and social science often lack cheap, clear, low-noise evaluators.

\subsection{Self-Driving Laboratories}

Self-driving laboratories come closest to a scientific closed loop. They are naturally expressed as Bayesian active learning or a partially observable Markov decision process. Given $b_t(\omega)=P(\omega\mid D_t)$, the next experiment is:
\begin{equation}
a_t=\argmax_{a\in A}
\left[
\E_{y\sim P(y\mid a,b_t)}U(y,a)
+\eta I(\omega;Y_a\mid D_t)
-\lambda \Cost(a)
-\kappa \Risk(a)
\right].
\end{equation}
After the experiment, $y_t\sim P(y\mid \doop(a_t),\omega^\ast)$ and:
\begin{equation}
b_{t+1}(\omega)=
\frac{P(y_t\mid \doop(a_t),\omega)b_t(\omega)}
{\int_{\Omega}P(y_t\mid \doop(a_t),\omega')b_t(\omega')d\omega'}.
\end{equation}

Boiko et al.'s Coscientist chemistry system demonstrates a GPT-4-driven workflow that combines internet and documentation search, code execution, and experimental automation to design, plan, and execute chemistry experiments, including palladium-catalyzed cross-coupling tasks \citep{Boiko2023Coscientist}. A review of self-driving laboratories describes SDLs as systems that combine automated experimental workflows with autonomous experimental planning and hold potential for accelerating chemistry and materials discovery \citep{Tom2024SDL}. A-Lab integrates computation, literature-derived historical data, machine learning, active learning, and robotic solid-state synthesis. After Nature's correction boundary, it should be described as realizing target compounds under corrected structural-characterization criteria, not as a settled discovery of 43 scientifically new materials \citep{Szymanski2023ALab,Szymanski2026ALabCorrection}.

The scientific view is:
\[
\boxed{\text{scientific knowledge} \approx \text{a stable regularity under intervention}.}
\]
This route is expensive and difficult to scale. Experiments cost far more than text generation; results are noisy, $y=f(a,\omega)+\epsilon$; and the action space is constrained by instruments, robots, safety, and protocols. The A-Lab controversy also shows that contact with the physical world does not remove interpretive risk. It increases the need for experimental audit, data standards, phase identification, novelty checks, and independent replication \citep{Chawla2026ALabCorrection}.

\section{Comparison Across Routes}

\begin{table}[htbp]
\centering
\small
\caption{Epistemological comparison of AI-driven science routes and artifact frameworks}
\begin{tabularx}{\textwidth}{@{}p{2.9cm}Y Y Y Y@{}}
\toprule
Route & Minimal knowledge unit & Mathematical form & View of science & Core risk \\
\midrule
Specialized scientific model & Prediction & $p_\theta(y\mid x)$ & Science as prediction of natural structure & Prediction mistaken for explanation, validation, or licensed discovery \\
LLM research assistant & Textual hypothesis & $P_\phi(h\mid \Know,q)$ & Science as semantic recombination & Plausibility mistaken for truth \\
Multi-agent co-scientist & Testable hypothesis & $(h,\mathcal{T}(h),s(h))$ & Science as hypothesis competition and critique & Experimental actionability mistaken for general scientific discovery \\
AI Scientist pipeline & Research artifact & $z=(h,c,e,r,p)$ & Science as executable research workflow & Artifact completeness mistaken for evidence-licensed contribution \\
Algorithmic/proof agent & Program or proof & $V(p)$ or $\operatorname{Check}(\pi)=1$ & Science as verifiable construction & Limited to evaluator-rich domains \\
Self-driving laboratory & Interventional regularity & $P(Y\mid \doop(a),h,c)$ & Science as stable response under intervention & World grounding mistaken for sufficient explanation or generalization \\
Typed artifact framework & Residual artifact & $R_{\mathrm{CSC}}(A')$ & Science as regime-aware artifact transition & Provenance completeness mistaken for claim calibration \\
\bottomrule
\end{tabularx}
\end{table}

The table shows that these routes do not solve the same problem. Specialized predictive models place meaning in output distributions. LLMs place meaning in corpus relations. Co-scientist systems place meaning in a hypothesis competition network. AI Scientist systems place meaning in executable research artifacts. AlphaEvolve and formal proof systems place meaning in objects accepted by evaluators. Self-driving laboratories place meaning in experimentally observed interventions. Typed artifact frameworks place meaning in provenance-preserving transitions between regimes. Evidential semantics asks a different but complementary question: which of these outputs licenses which scientific claim?

The shared failure mode is claim drift: the language, framing, or social interpretation of a result migrates toward a stronger claim while the evidence profile has not comparably strengthened. Let $\operatorname{Drift}_{\Dom,V}(C_{\mathrm{literal}},C_{\mathrm{implied}},E)>0$ denote the case in which the implied claim is stronger than the literal or licensed claim and is not contained in $\mathcal{L}_{\Dom,V}(E)$. The drift takes different forms in different routes:

\begin{table}[H]
\centering
\small
\caption{Claim-drift modes across AI science routes}
\begin{tabularx}{\textwidth}{@{}p{3.1cm}p{4.2cm}Y@{}}
\toprule
Route & Drift mode & Unlicensed migration \\
\midrule
Specialized scientific model & Prediction-to-discovery drift & A predicted structure, property, or stable candidate is described as a validated discovery. \\
LLM research assistant & Plausibility-to-truth drift & A fluent hypothesis or proof sketch is treated as true without independent adjudication. \\
Multi-agent co-scientist & Actionability-to-discovery drift & An experimentally actionable candidate is redescribed as a general scientific or translational discovery. \\
AI Scientist pipeline & Artifact-to-knowledge drift & A complete paper, codebase, or review artifact is treated as established scientific contribution. \\
Algorithmic/proof agent & Formalization-to-world drift & A checked result inside a formal system is overextended to domains not captured by the formalization. \\
Self-driving laboratory & Intervention-to-generalization drift & A real experimental response is redescribed as a general law or application-ready result. \\
Typed artifact framework & Provenance-to-warrant drift & Preserved provenance or schema transition is treated as evidence that a strong claim is licensed. \\
\bottomrule
\end{tabularx}
\end{table}

We can therefore distinguish four epistemic stages. They are not all forms of cognitive understanding in the same sense; rather, they are increasing levels of warrant for scientific assertion:
\begin{align}
W_{\mathrm{text}}(h)&=P_\phi(\mathrm{continuation}\mid h,\Know),\\
W_{\mathrm{conseq}}(h)&=\mathbf{1}\{\Conseq_{\Dom}(h)\ \text{is specified and testable in }\Dom\},\\
W_{\mathrm{adjudicated}}(h,E,V)&=\mathbf{1}\{\text{derived consequences of }h\text{ survive }V\text{ in }E\},\\
W_{\mathrm{license}}(C,E,V)&=\mathbf{1}\{C\in\mathcal{L}_{\Dom,V}(E)\}.
\end{align}
Here $W_{\mathrm{license}}$ is deliberately written as a membership test: a claim counts as evidence-licensed only if it belongs to the set of claims that the evidence permits. The consequence and adjudication warrants are also written schematically because their concrete forms are domain-specific: predictive warrant and causal warrant are important special cases in experimental and interventional sciences. This warrant notation also avoids confusing these stages with the belief-update operator $U$ introduced later.
These objects are not all scalars of the same type. The following relation is therefore a qualitative hierarchy of warrant, not a numerical inequality:
\[
W_{\mathrm{text}}\prec_{\mathrm{warrant}}
W_{\mathrm{conseq}}\prec_{\mathrm{warrant}}
W_{\mathrm{adjudicated}}\prec_{\mathrm{warrant}}
W_{\mathrm{license}}.
\]
This does not make linguistic plausibility useless. It means that its epistemic warrant is weaker. A scientific system must move textual hypotheses into consequence-bearing, externally adjudicated, and claim-calibrated structure.

\section{Adjudication and Claim Calibration}

The most important comparison is not model size but feedback quality. Let $V$ be the evaluator. Here $V$ means the system, test, proof checker, benchmark, experiment, or review process that judges a candidate; it should not be confused with $\mathcal{V}(h)$, the hypothesis-specific verification rule in the semantic tuple. Then:
\begin{equation}
\mathrm{EpistemicStrength}(V)\propto
\mathrm{Independence}(V)\times
\mathrm{Reliability}(V)\times
\mathrm{WorldGrounding}(V).
\end{equation}
The weakness of LLM self-evaluation is that $V_{\mathrm{LLM}}$ is highly correlated with $G_{\mathrm{LLM}}$. Its independence is low. Proof checkers, program tests, experiments, and statistical tests are more independent and therefore provide stronger epistemic warrant. But strong evaluation is still not enough. A validation result becomes scientifically meaningful only after it is converted into a calibrated claim:
\begin{equation}
\mathrm{ClaimQuality}(C,E,V)\propto
\mathrm{EpistemicStrength}(V)\cdot
\exp[-\lambda\,\Debt^\ell_{\Dom,V}(C,E)].
\end{equation}
Thus, the final epistemic bottleneck is not only evaluator strength, but claim calibration after evaluation. For a sound calibrated output $C^\ast\in\Cal_{\Dom}(C,E,V)$, membership in $\mathcal{L}_{\Dom,V}(E)$ is already guaranteed by definition; the quality penalty is therefore applied to the raw or intended claim before calibration.

\begin{table}[htbp]
\centering
\small
\caption{Evaluator strength across AI science routes}
\begin{tabularx}{\textwidth}{@{}p{3.2cm}Y p{2.8cm}@{}}
\toprule
Route & Main evaluator & Evaluator strength \\
\midrule
LLM research assistant & Language probability and human expert review & Weak to medium \\
Multi-agent co-scientist & Multi-agent critique, human filtering, later experiment & Medium, depending on external validation \\
AI Scientist pipeline & Benchmark, automatic review, paper structure, human review & Medium, but Goodhart-prone \\
AlphaFold/GNoME & Structure, energy, stability prediction plus experiment or computation & Medium to strong \\
AlphaEvolve/proof agent & Automated evaluator, program test, proof checker & Strong, but domain-limited \\
Self-driving laboratory & Real experimental result, statistics, replication & Strong, but costly and noisy \\
\bottomrule
\end{tabularx}
\end{table}

Reliability does not increase monotonically with automation. End-to-end automation can optimize the wrong proxy. Formal systems may be less general but more trustworthy because their evaluators are strict. Self-driving laboratories touch the world but require heavy engineering, statistical, and replication governance. Across all routes, the final claim should be set by the pair $(V,\Cal_{\Dom})$: the evaluator determines what survived, and the calibration operator determines what may be asserted.

\section{Toward an AI-Augmented Calibrated Scientific Loop}

All AI science systems can be decomposed into five operators. These operators answer five different questions: what to propose, what follows from it, whether it survives a test, how beliefs change, and what may finally be claimed.

\begin{table}[htbp]
\centering
\small
\caption{Five operators in the calibrated AI science loop}
\begin{tabularx}{\textwidth}{@{}p{1.6cm}p{3.4cm}p{3.4cm}Y@{}}
\toprule
Operator & Input & Output & Question answered \\
\midrule
$G$ & Knowledge $\Know$, accumulated data $D_t$, question $q$ & Hypotheses $H$ & What candidates should be proposed? \\
$M$ & Hypothesis $h$, action $a$, context $x$ & Consequence $\chi_h(a,x)\in\mathcal{Q}_{\Dom}$ & What observable, formal, diagnostic, or evidential consequence follows? \\
$V$ & Hypothesis $h$, action $a$, outcome $y$ & Score, pass/fail result, or test statistic & Did the candidate survive an external check? \\
$U$ & Belief state $B_t$, action $a_t$, outcome $y_t$ & Updated belief $B_{t+1}$ & How should the system revise its beliefs? \\
$\Cal_{\Dom}$ & Raw claim $C$, evidence $E$, validation rule $V$, domain context $\Dom$ & Calibrated claim $C^\ast$ or maximal licensed frontier & What assertion right has the system acquired? \\
\bottomrule
\end{tabularx}
\end{table}

In formula form, the same five-operator loop is:
\begin{align}
G &: (\Know,D_t,q)\rightarrow H,\\
M_{\Dom} &: (h,a,x)\rightarrow \chi_h(a,x)\in\mathcal{Q}_{\Dom},\\
V &: (h,a,y)\rightarrow \{0,1\}\ \text{or}\ \mathbb{R},\\
U &: (B_t,a_t,y_t)\rightarrow B_{t+1},\\
\Cal_{\Dom} &: (C,E,V)\rightarrow \Max_{\preceq_{\Dom}}(\downarrow C\cap\mathcal{L}_{\Dom,V}(E)).
\end{align}
In interventional or probabilistic settings, the local consequence may take the familiar form $\chi_h(a,x)=P(Y\mid h,\doop(a),x)$. In the general case, $M$ is not restricted to future-oriented numerical prediction. It may derive retrodictions, diagnostic features, proof obligations, classification constraints, measurement signatures, or expected traces. $G$ generates hypotheses. $M$ turns hypotheses into observable, formal, or measurement-level consequences. $V$ adjudicates whether a hypothesis survives. $U$ updates beliefs. $\Cal_{\Dom}$ calibrates the final claim to the evidence. The ideal system is therefore:
\begin{equation}
\boxed{\mathrm{AI\ Scientific\ Exploration}
=(G+M+V+U)+\Cal_{\Dom}.}
\end{equation}
Equivalently, the first four operators form the evidence-producing loop, while $\Cal_{\Dom}$ manages assertion rights:
\begin{equation}
\boxed{
\mathrm{AI\ Scientific\ Exploration}
=
\underbrace{G+M+V+U}_{\text{evidence-producing loop}}
+
\underbrace{\Cal_{\Dom}}_{\text{assertion-right management}}.
}
\end{equation}
The mature form of AI science may therefore not be a machine that autonomously writes papers, but a system that manages assertion rights: it proposes hypotheses, exposes them to adjudication, records evidence, and outputs only those claims for which it has acquired warrant.

This also defines a route-level failure pressure. Let:
\begin{equation}
\Shear_r(t)=
\frac{
G^{\mathrm{norm}}_r(t)+M^{\mathrm{norm}}_r(t)
}{
V^{\mathrm{norm}}_r(t)+Q^{\mathrm{cal,norm}}_{r,\Dom}(t)+\epsilon
}.
\end{equation}
When $\Shear_r(t)\gg 1$, the route can generate or predict more candidates than it can adjudicate and calibrate. The expected failure modes are hypothesis inflation, artifact inflation, overclaiming, and trust collapse. Foundation-model progress can improve $G$ and $M$ faster than it improves $V$ and $\Cal_{\Dom}$. Stronger models can therefore increase epistemic risk unless adjudication and calibration scale with them.

The ratio is not proposed as a directly measured universal quantity. It is a diagnostic abstraction for simulation and system design: all four capacities are normalized indices on a common synthetic scale, and $\epsilon>0$ prevents division by zero. The diagnostic asks whether the production side of a route is scaling faster than the adjudication and assertion-right side.

A more appropriate discovery objective is:
\begin{equation}
\begin{aligned}
W(h,C_h,E_h)
&=\alpha\log
\left(
\frac{P(D_{\mathrm{test}}\mid h)}
{P(D_{\mathrm{test}}\mid h_0)}
\right)\\
&\quad+\beta I(\omega;Y_{\mathcal{T}(h)})
+\gamma\mathrm{CausalScope}(h)\\
&\quad+\delta\mathrm{Novelty}(h)
-\lambda\mathrm{Complexity}(h)
-\mu\mathrm{Cost}(h)\\
&\quad-\nu\Debt^\ell_{\Dom,V}(C_h,E_h).
\end{aligned}
\end{equation}
Here $C_h$ is the claim the system intends to make about hypothesis $h$, and $E_h$ is the evidence collected for that hypothesis. This is a better target than simply maximizing paper score or LLM plausibility because it penalizes claims that exceed evidence.

In practice, a robust pipeline should:
\begin{enumerate}
  \item generate hypotheses with an LLM or co-scientist system, $h_t\sim P_{\mathrm{LLM}}(h\mid \Know,D_t,q)$;
  \item derive domain-appropriate testable consequences using a model, simulator, theory, measurement framework, or formal system:
  \[
    \chi_h(a,x)\in\mathcal{Q}_{\Dom};
  \]
  In interventional experimental settings, this may specialize to $\chi_h(a,x)=P_h(Y\mid \doop(a),x)$;
  \item select the highest information-per-cost test:
  \[
    a_t=\argmax_a I(h;Y_a\mid D_t)/\Cost(a);
  \]
  \item adjudicate with experiment, simulation, proof checker, benchmark, or statistical test:
  \[
    V(h,a_t,y_t);
  \]
  \item update beliefs:
  \[
    P(h\mid D_{t+1})\propto P(y_t\mid h,a_t,x_t,D_t)P(h\mid D_t).
  \]
  \item output the strongest evidence-licensed claim or maximal licensed claim frontier:
  \[
    C^\ast\in\Cal_{\Dom}(C,E,V).
  \]
\end{enumerate}
Without the sixth step, an AI science loop can still produce overstrong language. A system may generate a hypothesis, derive a testable consequence, run an experiment or proof check, and update a belief while still reporting a claim that exceeds the evidence. The goal is therefore not to replace science with a single autonomous AI scientist. It is to build an AI-augmented calibrated scientific loop: domain models derive consequences, LLM agents expand the hypothesis space, automated platforms execute, formal and experimental systems adjudicate, $\Cal_{\Dom}$ calibrates the claim, and human scientists define the problem, boundaries, and final interpretation.

\section{Illustrative Synthetic Dynamics}

AISim-Cal is introduced here as a computational thought experiment and synthetic sensitivity model for the calibration semantics developed above. Its purpose is not to forecast the future productivity of AI science, to rank institutions, or to predict which route will dominate a field. The simulation is included to illustrate the internal behavior of the proposed concepts, not to validate the framework or the superiority of any route. Rather, it turns the paper's conceptual variables into a stylized dynamical system so that assumptions about generation, evaluation, calibration, cost, and claim strength can be inspected explicitly. In this sense, the simulation is closer to an epistemological model of possible mechanism relations than to an empirical measurement program \citep{Winsberg2010Simulation}. All numerical quantities in this section are dimensionless synthetic quantities under a chosen parameterization, and the reported values should be read as illustrative methodological diagnostics rather than route performance measurements.

The model represents each route as a policy-like composition of the same operators used in the calibrated loop. A route has generation capacity $G_r(t)$, consequence-modeling capacity $M_r(t)$, evaluator strength $V_r(t)$, belief-update capacity $U_r(t)$, and route-level calibration quality $Q^{\mathrm{cal}}_{r,\Dom}(t)$. The symbol $Q^{\mathrm{cal}}_{r,\Dom}(t)$ is a scalar simulation parameter; it is kept distinct from the set-valued calibration operator $\Cal_{\Dom}$. The simulated state variables include route capability, evaluator strength, calibration quality, cost burden, update-loop maturity, goal clarity, true licensed claim production, final claim level, licensed claim level, and overclaim burden. The seven route families are specialized foundation models, human-led LLM assistants, multi-agent co-scientist systems, end-to-end AI Scientist pipelines, algorithmic and mathematical discovery agents, self-driving laboratories, and a \emph{Hybrid-Cal} loop. The last item is not treated as an empirically established route already proven superior. It is a design hypothesis: a deliberately calibrated architecture that combines broad hypothesis generation with stronger external evaluators, explicit update loops, and conservative claim calibration. The default run uses 128 Monte Carlo draws, random seed 4317, and a 2026--2035 synthetic horizon.

\begin{table}[htbp]
\centering
\scriptsize
\caption{AISim-Cal default route parameters used for the illustrative base setting}
\begin{tabularx}{\textwidth}{@{}p{2.75cm}*{8}{>{\centering\arraybackslash}X}@{}}
\toprule
Route & $G$ & $M$ & $V$ & $U$ & $Q^{\mathrm{cal}}$ & Cost & Throughput & Overclaim \\
\midrule
Specialized foundation model & 0.52 & 0.86 & 0.62 & 0.48 & 0.63 & 0.56 & 0.76 & 0.11 \\
Human-led LLM assistant & 0.70 & 0.50 & 0.48 & 0.45 & 0.54 & 0.34 & 0.66 & 0.18 \\
Multi-agent co-scientist & 0.80 & 0.58 & 0.54 & 0.62 & 0.57 & 0.48 & 0.78 & 0.20 \\
End-to-end AI Scientist & 0.82 & 0.56 & 0.45 & 0.58 & 0.43 & 0.43 & 0.92 & 0.28 \\
Algorithmic/math agent & 0.72 & 0.62 & 0.88 & 0.78 & 0.76 & 0.30 & 0.70 & 0.07 \\
Self-driving laboratory & 0.66 & 0.64 & 0.80 & 0.82 & 0.69 & 0.92 & 0.52 & 0.10 \\
Hybrid-Cal loop & 0.72 & 0.68 & 0.76 & 0.74 & 0.82 & 0.58 & 0.68 & 0.06 \\
\bottomrule
\end{tabularx}
\end{table}

The table reports dimensionless design parameters, not measurements. The route ordering in later figures is therefore a consequence of this illustrative parameterization and must not be interpreted as an empirical ranking of real systems.

AISim-Cal also includes a domain-level goal-clarity parameter $\Gamma_{\Dom}\in[0,1]$. This parameter records whether the target of inquiry is sufficiently specified to support a meaningful comparison of routes. High $\Gamma_{\Dom}$ corresponds to settings in which a domain has a clear objective, scoring rule, admissible evidence type, and claim-value target; examples include formal proof, benchmarked algorithmic improvement, or a well-defined screening objective. Low $\Gamma_{\Dom}$ corresponds to open-ended problem formulation, where the central task is still to decide what should count as a good question, a meaningful consequence relation, a valid evaluator, or a licensed claim. AISim-Cal uses a threshold $\tau_\Gamma=0.45$: when $\Gamma_{\Dom}<\tau_\Gamma$, the model still reports diagnostic scores but does not license a route ordering. It marks the corresponding row as \emph{no licensed ordering}. The threshold is illustrative rather than empirically estimated, and the companion simulation exposes it as a configurable parameter. In that regime, route comparison moves upstream to problem definition and evaluator construction rather than downstream to route selection.

In addition to route-level parameters and goal clarity, AISim-Cal applies a route-domain applicability factor $A_{r,\Dom}\in[0,1]$. A route with strong evaluators is not automatically applicable to every scientific domain. Formal agents can dominate evaluator-rich formal domains, but they should not be interpreted as generally superior in wet-lab biology, chemical synthesis, or materials characterization unless their operators are domain-compatible. The reported utility is therefore computed as
\begin{equation}
\mathrm{LicensedUtility}_{r,\Dom}
=
A_{r,\Dom}\cdot
\mathrm{RawLicensedUtility}_{r,\Dom},
\end{equation}
where $\mathrm{RawLicensedUtility}_{r,\Dom}$ is the synthetic utility before route-domain coverage is applied. This additional factor makes the heatmap a conditional route comparison rather than a claim that evaluator strictness alone determines scientific usefulness.

\begin{table}[htbp]
\centering
\scriptsize
\caption{AISim-Cal route-domain applicability matrix $A_{r,\Dom}$ used in the illustrative run}
\begin{tabularx}{\textwidth}{@{}p{2.7cm}*{6}{>{\centering\arraybackslash}X}@{}}
\toprule
Route & Biology & Chemistry & ML & Materials & Math/alg. & Open-ended \\
\midrule
Specialized foundation model & 0.82 & 0.74 & 0.62 & 0.88 & 0.42 & 0.45 \\
Human-led LLM assistant & 0.78 & 0.62 & 0.72 & 0.60 & 0.60 & 0.72 \\
Multi-agent co-scientist & 0.90 & 0.72 & 0.66 & 0.72 & 0.55 & 0.66 \\
End-to-end AI Scientist & 0.60 & 0.50 & 0.86 & 0.56 & 0.70 & 0.58 \\
Algorithmic/math agent & 0.22 & 0.25 & 0.78 & 0.34 & 0.96 & 0.35 \\
Self-driving laboratory & 0.84 & 0.94 & 0.20 & 0.92 & 0.05 & 0.30 \\
Hybrid-Cal loop & 0.86 & 0.84 & 0.82 & 0.86 & 0.82 & 0.60 \\
\bottomrule
\end{tabularx}
\end{table}

AISim-Cal decomposes evaluator strength into three multiplicative components:
\begin{equation}
\mathrm{EvaluatorStrength}_r(t)=
\left[
\mathrm{Independence}_r(t)\times
\mathrm{Reliability}_r(t)\times
\mathrm{Grounding}_r(t)
\right]^{1/3}.
\end{equation}
Independence records how far the evaluator is separated from the generator. Reliability records whether repeated evaluation would give stable and auditable outcomes. Grounding records whether the evaluator is connected to a domain-relevant constraint, such as an experiment, proof checker, benchmark, simulator, statistical test, or structured external review. The geometric mean makes the evaluator bottleneck visible: a route cannot obtain high evaluator strength by excelling in only one component while the others remain weak. This decomposition preserves a central claim of the manuscript: a fluent internal critic can increase review-like structure without supplying the same epistemic warrant as an evaluator that is independent, reliable, and grounded in the target domain.

The simulated claim ladder follows the manuscript's L0--L6 scale. L0 denotes hypothesis generation. L1 denotes computational or model support. L2 denotes internal experimental support. L3 denotes external replication or independent support. L4 denotes causal mechanism support. L5 denotes generalizable scientific knowledge within stated boundaries. L6 denotes translational, application-ready, or otherwise high-stakes assertion. AISim-Cal then applies three calibration modes. In the simulation implementation, evaluator and evidence-side variables determine the licensed claim level, while calibration quality controls how strongly the final asserted claim is pulled back toward that licensed frontier. In the no-calibration mode, the final claim can exceed the licensed claim level. In the imperfect-calibration mode, claims are partially pulled back toward the licensed frontier but may still overshoot when evaluator strength or calibration quality is weak. In the perfect-calibration mode, the final claim is forced not to exceed the licensed claim level. Perfect calibration is included as a limiting reference condition, not as a realistic assumption about current AI science systems.

Four diagnostics summarize each run. The synthetic licensed-utility diagnostic measures useful claim production after the claim has been restricted to the licensed frontier and adjusted by route-domain applicability $A_{r,\Dom}$. The overclaim gap measures how far the final asserted level exceeds the licensed level. False-discovery burden combines unsupported claim production with overclaiming pressure. The evidence bottleneck index measures the mismatch between generation/modeling capacity and the combined strength of evaluation, updating, and calibration. These metrics are intentionally formal rather than empirical. They make visible the paper's claim that a route can look productive when judged by raw outputs while being weak when judged by evidence-licensed assertion.

The score-landscape diagnostic has two additional preconditions. First, a route comparison can be interpreted as an ordering only relative to a specified objective and scoring rule. If the domain is still asking what the right objective should be, then a numerical score is useful as an exploratory diagnostic but should not be promoted into a dominance statement. Second, the route must have sufficient domain applicability $A_{r,\Dom}$ for the comparison to be meaningful. For this reason, the score landscape still reports utility values for low-$\Gamma_{\Dom}$ settings, but it marks them as unordered rather than assigning route dominance under an arbitrary scalarization. It also treats strict formal evaluators as strong primarily in domains where the route itself is applicable. The full route--domain--scenario landscape is provided in Appendix~\ref{app:score-landscape}; the main text emphasizes the route utility and calibration-ablation diagnostics.

\begin{figure}[htbp]
\centering
\includegraphics[width=0.92\textwidth]{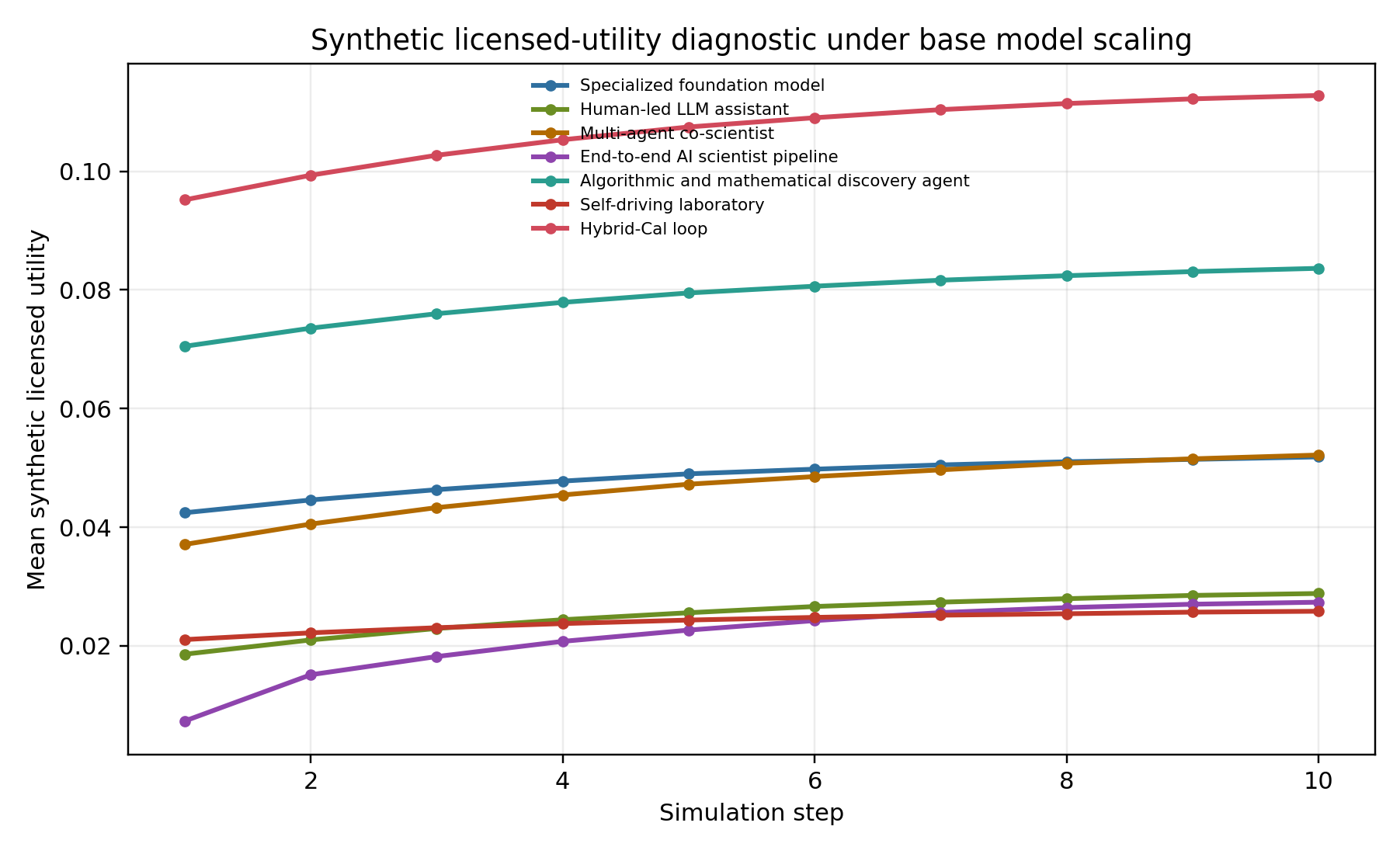}
\caption{Synthetic licensed-utility diagnostic across AISim-Cal routes. All quantities are dimensionless synthetic quantities and not empirical forecasts. Lines show means over the 128 Monte Carlo draws in the illustrative parameterization; propagated standard-error bands are small in this run and are provided in the companion outputs. The curves include goal clarity and route-domain applicability factors; they are conditional diagnostics of the calibration framework, not predictions of future scientific output. The ordering of routes is not a claim about real-world dominance; it is a consequence of the illustrative parameter setting.}
\label{fig:aisim-route-utility}
\end{figure}

Under this parameterization, the route-level outputs illustrate the difference between apparent productivity and a synthetic licensed-utility diagnostic. Routes with strict evaluators, especially the algorithmic and mathematical discovery route, obtain high licensed utility where the synthetic evaluator is comparatively independent, reliable, cheap to apply, and domain-compatible. The Hybrid-Cal loop also ranks highly in several scenarios because its parameters encode strong calibration quality, evaluator integration, and broader route-domain coverage. This is a conditional consequence of the design assumptions, not evidence that such a route has already been demonstrated. No inference should be drawn about real-world superiority of Hybrid-Cal or any route. Conversely, routes with high throughput but weaker calibration can accumulate larger overclaim burdens, especially when automatic artifact production or benchmark optimization is not matched by independent adjudication and claim weakening.

\begin{figure}[htbp]
\centering
\includegraphics[width=0.92\textwidth]{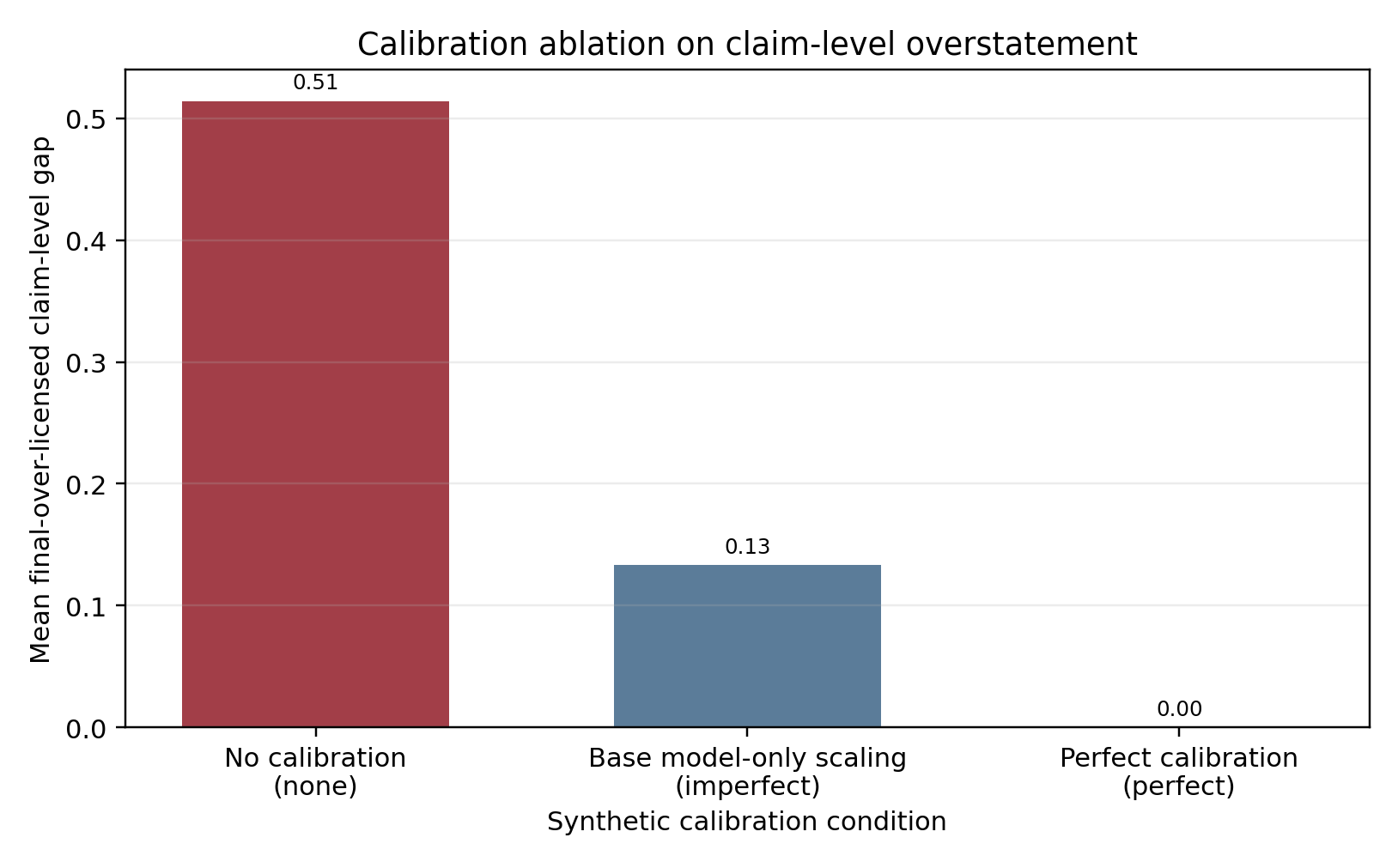}
\caption{Calibration ablation in AISim-Cal. All quantities are dimensionless synthetic quantities and not empirical forecasts. The no-calibration, imperfect-calibration, and perfect-calibration conditions show how the synthetic overclaim gap changes when the same route dynamics are passed through different claim-calibration regimes.}
\label{fig:aisim-calibration-ablation}
\end{figure}

The calibration ablation is the most direct synthetic illustration of the manuscript's core semantics. In the generated outputs, the mean final-over-licensed claim-level gap is largest when calibration is absent, smaller under imperfect calibration, and zero under the perfect-calibration reference condition. This pattern is stylized and conditional, but it expresses the formal point of $\Cal_{\Dom}$: external validation alone does not determine the permissible claim level. A system must still translate evidence into a licensed claim, and a weaker calibration operator leaves residual epistemic debt even when some validation exists.

The companion source repository contains CSV summaries, result notes, and generated figures for the AISim-Cal illustration. These artifacts are included to make the synthetic parameterization and figure generation auditable; they are not new empirical experimental data.

\FloatBarrier

\section{AI-System Design Checklist}

The framework above can be translated into a concrete design checklist for evaluating AI scientific exploration systems. The checklist does not require every system to contain all modules at the same level of maturity. A language-model assistant, a theorem-proving system, and a robotic laboratory will have very different engineering constraints. The purpose is instead to make the epistemic contract explicit. A valid claim of AI scientific exploration or discovery should specify the hypothesis object, consequence object, evaluator, update rule, and calibrated claim.

\begin{table}[H]
\centering
\scriptsize
\caption{Checklist for evaluating AI scientific exploration systems}
\begin{tabularx}{\textwidth}{@{}p{3.1cm}Y Y@{}}
\toprule
Design question & What should be specified & Failure mode if omitted \\
\midrule
What is generated? & The hypothesis $h$, candidate object, proof sketch, experiment plan, molecule, material, or algorithm produced by the system. & Fluency or artifact production is mistaken for scientific content. \\
What follows? & The consequence $\chi_h(a,x)$: observable response, retrodictive trace, benchmark outcome, proof obligation, assay response, diagnostic feature, measurement signature, or material property expected under the hypothesis. & The system proposes ideas that cannot be tested or compared. \\
Who evaluates it? & The evaluator $V$: proof checker, unit test, benchmark, simulator, wet-lab assay, statistical test, external cohort, expert review, or independent laboratory. & Internal self-critique is mistaken for external adjudication. \\
How independent is the evaluator? & Whether $V$ is independent from the generator $G$, whether data leakage is controlled, and whether the evaluation was designed before or after seeing candidates. & The system appears successful because the generator and evaluator share biases. \\
What evidence was obtained? & The evidence body $E$: result files, datasets, assay readouts, confidence intervals, ablations, controls, failed cases, or replication records. & A positive narrative is reported without auditable evidential objects. \\
What claim is made? & The explicit claim $C=\Claim(h,s,b,m)$, including its strength, boundary, and modal status. & The paper shifts from a modest result to an overstrong conclusion. \\
How is the claim calibrated? & The selected calibrated claim $C^\ast\in\Cal_{\Dom}(C,E,V)$ and the reason why $C^\ast$ belongs to $\mathcal{L}_{\Dom,V}(E)$. & Validation is treated as a license for any stronger downstream assertion. \\
What epistemic debt remains? & The residual $\Debt^\ell_{\Dom,V}(C,E)$, if the raw or implied claim exceeds the licensed set. & The paper hides an unpaid warrant deficit behind polished language. \\
What are the boundaries? & The contexts, populations, assays, datasets, domains, and deployment conditions under which the claim is intended to hold. & The evidential scope remains ambiguous: local, general, causal, or translational claims are not distinguished. \\
\bottomrule
\end{tabularx}
\end{table}
\FloatBarrier

This checklist also clarifies the evidential structure of AI science claims. A system can be impressive at $G$ but weak at $V$: it may generate creative hypotheses but lack independent validation. A system can be strong at $V$ but weak at $\Cal_{\Dom}$: it may obtain a valid benchmark result yet describe it with language that implies a broader scientific discovery. A system can be strong at $M$ but weak at $U$: it may derive accurate consequence profiles without updating beliefs about mechanisms. These distinctions are operational rather than merely terminological. They are the difference between a useful AI research tool, an auditable scientific discovery engine, and a paper-writing system that produces plausible but under-validated claims.

The checklist is intentionally compatible with different AI research cultures. In machine learning, $V$ may be a benchmark suite, ablation protocol, or held-out dataset. In formal mathematics, $V$ may be a proof assistant or proof checker. In materials science, $V$ may be a density-functional-theory calculation, phase-stability criterion, synthesis attempt, or structural characterization. In biology, $V$ may be an in vitro assay, perturbation experiment, animal model, independent cohort, or clinical study. The notation is shared, but the evidential threshold is domain-specific. This is why the calibration operator depends not only on $C$ and $E$, but also on $V$ and $\Dom$: the same evidence can license different claims in different domains.

For manuscript reporting, the framework suggests a simple discipline. A manuscript should state the raw AI output, the testable consequence derived from that output, the external or semi-external evaluator, the evidence actually obtained, the claim level on the claim ladder, and the reason stronger claims are not yet licensed. This discipline does not weaken the contribution. It makes evidential strengths and remaining validation needs explicit.

\section{Practical Implications}

For data-rich experimental domains, including life science, chemistry, materials science, and AI for biology, a stable configuration is to use specialized models to generate candidates, LLM or multi-agent systems for literature integration and mechanistic hypotheses, and then strict verification through database back-testing, negative controls, statistical tests, wet-lab experiments, or independent cohorts. The final claim should be reported at the correct level of the claim ladder: computational candidate, in vitro support, independent replication, causal mechanism, or translational claim. AI may generate proposals at scale, but scientists should retain control over problem definition, evaluation criteria, and final claim boundaries.

For AI and machine learning research itself, stronger automation is more defensible because experiments are cheap and feedback is fast. AI Scientist systems, code agents, automatic baseline search, and benchmark evaluation can be useful. The danger is that automatic tuning plus automatic writing may be mistaken for theory. Baselines, statistical robustness, ablations, data leakage, and novelty require explicit audit, and benchmark improvements should be calibrated against the claim they are used to support.

For mathematics, algorithms, and theoretical computer science, a particularly auditable configuration is to use LLMs for conjectures or proof sketches and connect them to Lean, Coq, Isabelle, SAT/SMT solvers, program verifiers, or strong benchmarks. The reason is not that the route is universally superior, but that candidate objects can be judged by strict external evaluators and then reported within explicit proof, program, or benchmark boundaries.

For chemistry, materials, and robotic laboratories, the long-term direction is the self-driving experimental loop. Short-term claims must not underestimate instruments, protocols, samples, data standards, safety, and replication governance. The A-Lab correction is a warning: even when AI enters the physical world, humans must still audit experimental definitions, phase identification, database deduplication, and novelty claims. World contact strengthens evidence, but it does not by itself determine whether the licensed claim is synthesis success, new phase discovery, generalizable materials rule, or application-ready material.

\section{Conclusion}

AI scientific exploration is uneven across evaluator regimes. Specialized predictive models illustrate high consequence-derivation capacity without full autonomy. Research assistants illustrate broad hypothesis and artifact support under human responsibility. Multi-agent systems illustrate structured hypothesis search. End-to-end AI Scientist systems illustrate workflow automation under constrained computational settings. Self-driving laboratories illustrate world-grounded loops with high engineering and validation costs. Mathematics and algorithms illustrate the special epistemic value of strict evaluators. These are conditional contrasts about evidence and claim licensing, not forecasts about which route will dominate. Across all routes, the central issue is not only whether the system generates hypotheses or obtains validation, but whether it converts validation into an appropriately bounded scientific claim.

The framework can be summarized by three calibration laws. First, no claim without license:
\begin{equation}
C\ \text{is publishable as a scientific assertion only if}\ C\in\mathcal{L}_{\Dom,V}(E).
\end{equation}
Second, validation does not determine claim level:
\begin{equation}
E\Vdash_{\Dom,V} C_1,\quad C_1\preceq_{\Dom} C_2
\quad\not\Rightarrow\quad
E\Vdash_{\Dom,V} C_2.
\end{equation}
The same evidence may license a weak claim without licensing a stronger claim. Third, automation amplifies the need for calibration:
\begin{equation}
\Shear_r(t)=
\frac{
G^{\mathrm{norm}}_r(t)+M^{\mathrm{norm}}_r(t)
}{
V^{\mathrm{norm}}_r(t)+Q^{\mathrm{cal,norm}}_{r,\Dom}(t)+\epsilon
}
\uparrow
\quad\Rightarrow\quad
\mathrm{Risk}_{\mathrm{overclaim}}\uparrow.
\end{equation}
When generation and consequence-derivation capacities scale faster than adjudication and calibration, AI systems may become more productive while also accumulating more epistemic debt.

Overclaiming is epistemic debt: it must be paid by additional evidence or cancelled by weakening the claim, narrowing its boundary, changing its modal status, or revising the hypothesis object.

The compressed thesis is:
\[
\boxed{\begin{gathered}
\text{LLMs expand hypotheses, consequence models give structure,}\\
\text{experiments or proof checkers give adjudication,}\\
\text{and evidence calibration gives licensed claims.}
\end{gathered}}
\]
More strictly:
\[
\boxed{\begin{gathered}
\text{scientific semantics}=\text{linguistic hypothesis}\\
+\text{testable consequence}+\text{external validation rule}\\
+\text{claim boundary.}
\end{gathered}}
\]
The main bottleneck in AI science is not the ability to generate hypotheses. It is not only the transition from $P_{\mathrm{language}}(h)$ to $\Conseq_{\Dom}(h)$, with $P_{\mathrm{world}}(Y\mid \doop(a),h)$ as one interventional special case. It is also the transition from validation result to licensed assertion. An AI system may generate a plausible hypothesis, derive a testable consequence, and even obtain experimental support. It must still answer: what strength of claim does this evidence permit, and what epistemic debt remains? On this framework, a reliable AI scientist would not merely write papers or run experiments. It would manage scientific assertion rights by transforming candidate hypotheses into evidence-licensed, boundary-explicit, and strength-calibrated scientific claims.

The practical purpose of the framework is comparative. It does not claim that AI science has already converged on one architecture, nor that any current system has solved autonomous discovery in full. It proposes a language for comparing systems that are otherwise difficult to compare: LLM assistants, co-scientists, AI Scientist pipelines, formal proof agents, specialized scientific models, and self-driving laboratories. The comparison becomes sharper when every system is asked the same questions: what does it generate, what consequence does it derive, who evaluates it, how is belief updated, and what final claim is licensed by the evidence?

The same discipline also explains why minimal reconstruction is difficult and valuable. Claim calibration is not always a move toward narrower language. Sometimes it is a move toward a broader structural assertion that reconstructs a licensed common pattern from many heterogeneous local results. Such a claim is scientifically powerful precisely because it is rare: it must preserve the local evidence, expose the shared structure, and leave the uncompressed residuals visible rather than converting them into unsupported universality.

The practical unit of AI-assisted scientific reporting is therefore not a claim alone, but a licensed claim paired with an explicit non-claim.

\section*{Limitations}

This paper is a theoretical synthesis and methodological comparison. It does not introduce new empirical experimental data. The AISim-Cal section adds a synthetic simulation and generated figure/CSV artifacts for sensitivity-style illustration of the calibration framework; those outputs are not empirical forecasts and should not be interpreted as measurements of future scientific productivity. Several 2025--2026 systems are still changing quickly, and their results should be interpreted within the boundaries stated by the original papers or official reports. This is especially important for GPT-5 science case studies, AI Scientist workshop review results, AlphaEvolve infrastructure claims, and the A-Lab materials controversy. The cases are used here as epistemological comparators, not as proof that any route has reached mature autonomous science.

\section*{Data Availability}

No new empirical experimental data were generated. The accompanying clean preprint source files contain the AISim-Cal synthetic outputs, including CSV summaries and generated figures. External source caches used for citation checking are not part of the clean preprint source release; external materials should be accessed through the public papers, official pages, or news reports cited in the reference list.

\section*{Code Availability}

The AISim-Cal synthetic simulation scripts, configuration files, generated CSV summaries, and generated figures are maintained with the clean preprint source files. They are illustrative research artifacts for inspecting the calibration framework, not empirical forecasts of future scientific productivity.

\section*{Ethics Statement}

This paper does not involve human participants, animal experiments, or sensitive personal data.

\section*{Competing Interests}

The author declares no financial competing interests. The analysis is based on public sources, cited literature, and the evidential boundaries explicitly stated in the manuscript.

\section*{Funding Statement}

No dedicated funding was received for this paper.

\section*{Author Contributions}

Hongmin Li developed the conceptual framework, argument structure, and claim-calibration interpretation. AI tools assisted with English drafting, structural organization, LaTeX preparation, reference formatting, and compilation checks. The author is responsible for final accuracy, citation completeness, and academic integrity.

\section*{AI Use Statement}

AI tools were used for drafting assistance, structural organization, LaTeX formatting, bibliography preparation, and compilation checks. The author reviewed all generated text, verified the argument structure, and is responsible for all citations, mathematical definitions, interpretations, and final claims.

\clearpage
\appendix

\section{AISim-Cal Score Landscape}
\label{app:score-landscape}

\begin{center}
\centering
\makebox[\textwidth][c]{\includegraphics[width=1.20\textwidth]{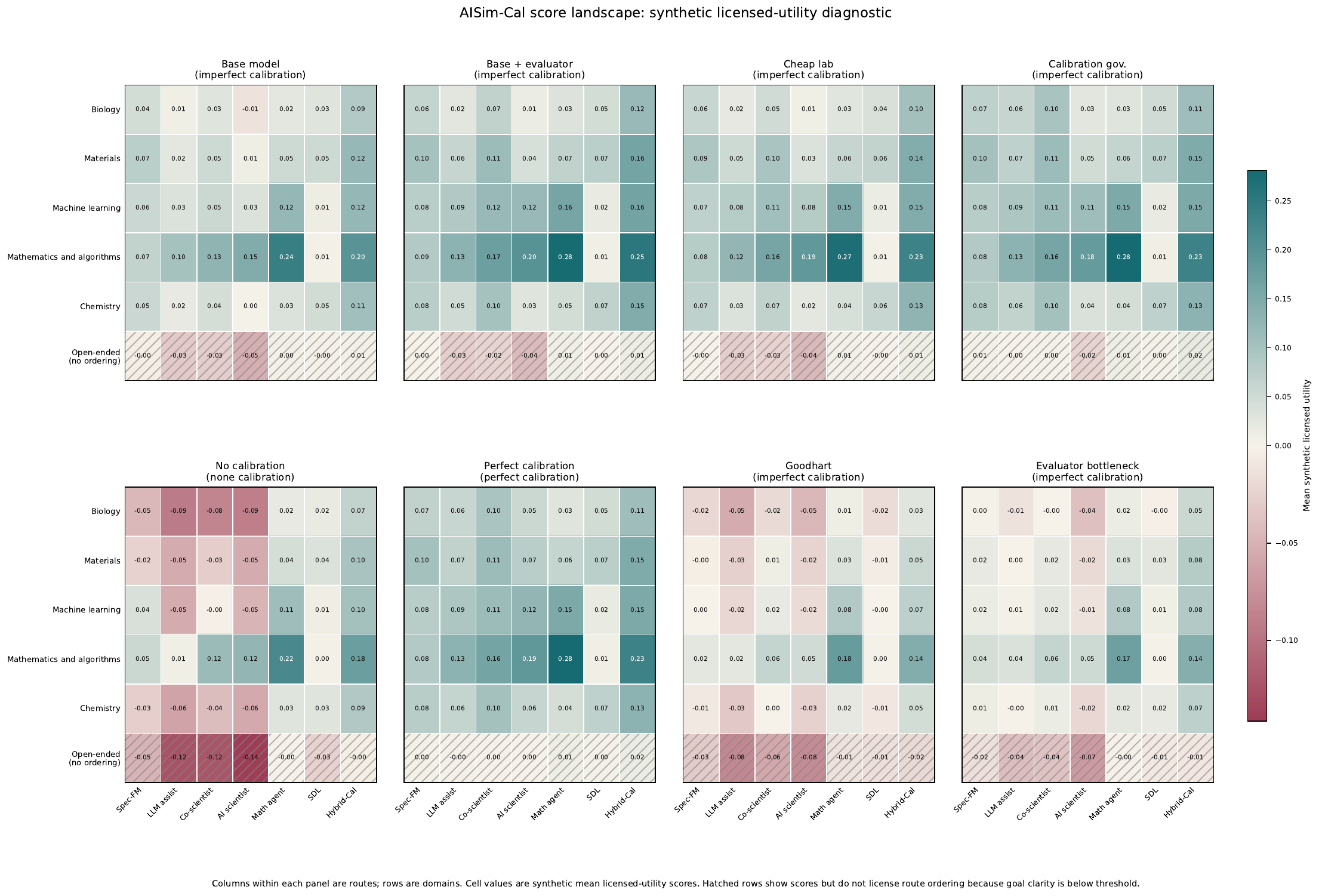}}
{\captionsetup{hypcap=false}
\captionof{figure}{AISim-Cal route--domain--scenario score landscape. Each panel corresponds to one scenario and calibration mode; each cell reports the synthetic licensed-utility diagnostic for one route in one domain. The values are dimensionless scores under the chosen synthetic parameterization, including goal clarity and route-domain applicability $A_{r,\Dom}$; they are not empirical forecasts. Hatched rows show low-goal-clarity regimes: the scores remain visible as exploratory diagnostics, but they do not license a route ordering because the scientific task is still problem formulation, evaluator construction, and claim-boundary definition.}
\label{fig:aisim-score-landscape}}
\end{center}

\newpage
\bibliography{references}

\end{document}